\documentclass{article}

\usepackage{arxiv}

\usepackage[utf8]{inputenc} 
\usepackage[T1]{fontenc}    
\usepackage{url}            
\usepackage{booktabs}       
\usepackage{amsfonts}       
\usepackage{nicefrac}       
\usepackage{microtype}      
\usepackage{lipsum}
\usepackage{graphicx}
\graphicspath{ {./images/} }

\usepackage[T1]{fontenc}

\usepackage{mathptmx}
\raggedright
\usepackage{standalone}
\usepackage{graphicx}
\usepackage{float}

\usepackage[algo2e,linesnumbered,ruled,vlined]{algorithm2e}
\usepackage{amsthm}
\usepackage{subcaption}
\usepackage{graphicx}
\usepackage{booktabs}
\usepackage{amssymb}
\usepackage{bm}
\usepackage{bbm}
\usepackage{enumitem}
\usepackage{amsmath}
\usepackage{booktabs}
\usepackage{array}
\usepackage{colortbl}
\usepackage{xcolor}
\usepackage{multirow}
\usepackage{makecell}
\theoremstyle{plain}
\newtheorem{theorem}{Theorem}

\usepackage{arydshln}

\usepackage[labelfont=bf]{caption}

\newtheorem*{lemma}{Lemma}
\usepackage{setspace}
\newcommand\independent{\protect\mathpalette{\protect\independenT}{\perp}}
\def\independenT#1#2{\mathrel{\rlap{$#1#2$}\mkern2mu{#1#2}}}
\usepackage{pgfplotstable}
\usepackage[flushleft]{threeparttable}
\usepackage[toc,title,page]{appendix}
\pgfplotsset{compat=1.18}

\usepackage{balance}       
\usepackage{graphics}      
\usepackage[T1]{fontenc}   
\usepackage{txfonts}
\usepackage{mathptmx}

\usepackage{algpseudocode}

\usepackage{authblk}
\usepackage[section]{placeins}

\usepackage[small,compact]{titlesec}

\providecommand{\keywords}[1]
{
  \small	
  \textbf{\textit{Keywords---}} #1
  \normalsize
}
\usepackage[pdfstartview=XYZ,
bookmarks=true,
colorlinks=true,
linkcolor=blue,
urlcolor=blue,
citecolor=blue,
pdftex,
bookmarks=true,
linktocpage=true, 
hyperindex=true
]{hyperref}

\usepackage{orcidlink}

\usepackage{booktabs} 
\usepackage{siunitx} 
\usepackage{amsmath} 

\title{using machine bias to measure human bias}

\author{
  Wanxue Dong \orcidlink{0000-0003-1668-7655}\\
  Department of Decisions, Operations and Technology\\
  The Chinese University of Hong Kong Business School\\
  \texttt{wanxuedong@cuhk.edu.hk}
  \and
  Maria De-Arteaga \orcidlink{0000-0003-2297-3308}\\
  Department of Information, Risk, and Operations Management\\
  McCombs School of Business\\
  The University of Texas at Austin\\
  \texttt{dearteaga@mccombs.utexas.edu}
  \and
  Maytal Saar-Tsechansky \orcidlink{0000-0002-2400-3762}\\
  Department of Information, Risk, and Operations Management\\
  McCombs School of Business\\
  The University of Texas at Austin\\
  \texttt{Maytal.Saar-Tsechansky@mccombs.utexas.edu}
}
\begin{document}
\maketitle

\begin{abstract}
Biased human decisions have consequential impacts across various domains, yielding unfair treatment of individuals and resulting in suboptimal outcomes for organizations and society. In recognition of this fact, organizations regularly design and deploy interventions aimed at mitigating these biases. However, \emph{measuring} human decision biases remains an important but elusive task. Organizations are frequently concerned with mistaken decisions disproportionately affecting one group. In practice, however, this is typically not possible to assess 
due to the scarcity of a gold standard: a label that indicates what the correct decision would have been.  
In this work, we propose a machine learning-based framework to assess bias in human-generated decisions when gold standard labels are scarce. We provide theoretical guarantees and empirical evidence demonstrating the superiority of our method over existing alternatives. This proposed methodology establishes a foundation for transparency in human decision-making, carrying substantial implications for managerial duties, and offering potential for alleviating algorithmic biases when human decisions are used as labels to train algorithms. 
\end{abstract}
\keywords{Human Labels,
Label Bias,
Crowdsourcing,
Bias Assessment,
Machine Learning,
Information Systems}

\section{Introduction}
\label{sec:intro}

Biased human assessments that result in suboptimal and unfair decisions are an important concern for organizations and society. In the realm of expert domains, for example, unequal treatment by medical professionals results in racial and ethnic disparities in healthcare \cite{nelson2002unequal}, and disparities also exist in the quality of healthcare provided to men and women \cite{jetten2019social,augoustinos2019language}. Similarly, it has been shown that decisions in recruiting and hiring can encode racial bias \cite{bertrand2004emily}. The problem also affects non-expert assessments. In the field of crowdsourcing, which has emerged as a promising avenue to collect large amounts of annotations required by data-hungry information systems \cite{lukyanenko2019expecting,hashim2024real,steelman2014data,wang2017cost}, 
biases constitute a central challenge threatening the usability of such data \cite{davani2023hate,saeed2022crowdsourced,chung2023ai}. For example, partisan biases can affect the quality of fact-checking efforts that rely on crowd annotators \cite{saeed2022crowdsourced,chung2023ai}.

\setlength{\parindent}{20pt}
Approaches to identify bias, and interventions to mitigate it, typically require a way to  \emph{measure} the possible presence of bias. 
For example, in the context of US equal opportunity employment law, the four-fifth rule indicates that if one group is selected at less than four-fifths the rate of another group, that is considered a presumption of adverse impact \cite{barocas2016big}. Similar measures are used in interventions aimed to mitigate decision biases. For example, the Rooney Rule, implemented by the National Football League (NFL) to promote the consideration of underrepresented candidates for head coach and senior positions, indicates that a certain rate or quota of minority candidates must be interviewed  \cite{collins2007tackling}. The approach, which has been credited with some positive outcomes \cite{collins2007tackling}, has been adopted by companies in efforts to improve diversity in hiring \cite{seetharaman2015facebook}. %

Such measures, which compare the \emph{rate} at which a certain decision is made for members of different groups, but do not consider the \emph{quality} of the decisions and the corresponding disparities in quality, are popular due to their ease of implementation but are fraught with limitations. Their ease of use comes from the fact that the only information necessary to compute them is (1) the decisions, which we refer to as $Y'$; and (2) the group membership, which we refer to as $A$. One does not require information regarding the correct decision, which we denote $Y$. For the same reason, however, these measures often fail to capture what we care about. If the base rates are different across groups, then always making the correct decision could yield different rates across groups \cite{chouldechova2017fair}. For example, due to sociocultural factors, the rate of people eligible for a position may be different across groups. Similarly, if we were interested in assessing bias in healthcare, if a disease is more frequent for one demographic group than others, then we would expect treatment for that disease to be prescribed more frequently to the group that has a higher incidence of the disease. Therefore, different rates do not always imply biased decisions. On the flipside, decisions that are biased may still yield equal rates. 
For instance, the Rooney Rule can be easily ``gamed" by choosing to interview less qualified candidates from the minority group. Furthermore, ill intent is not necessary to yield worse quality decisions for one group than the other, if decision-makers are less able to assess individuals from one group due to lack of domain or contextual knowledge.

It is therefore desirable to have measures of bias that account for the \emph{quality} of the decisions. This is reflected in the emerging literature on measures of algorithmic bias \cite{de2022algorithmic}, where metrics such as the disparity in false positive rates---which measures the disparity in false positive errors across groups---are increasingly popular \cite{hardt2016equality}. In the context of assessing algorithmic bias, these measures are computed by comparing the algorithmic predictions with the labels that the algorithm is trained to predict. However, measuring such biases in human decisions is rarely possible, due to the scarcity of gold standard labels, $Y$. Without these, it is unfeasible to contrast human judgments with what the correct decision should have been, and assess whether error rates differ across groups. In expert domains, acquiring gold standard decisions is often not impossible but very costly. For example, gold-standard assessments in healthcare are often acquired from costly expert panels that require significant investments, reserved for select clinical studies \cite{GORDON2009205}. Although data-driven machine learning (ML) predictive modeling can be used to estimate ground truth labels and compare them with human judgments to assess decision bias, its performance is often suboptimal due to two main issues: (1) the limited availability of ground truth data, and (2) the presence of bias across all noisy labels, which skews the predictions and significantly compromises model quality. Our empirical evaluation adopted these methods as benchmarks and confirmed this hypothesis. A similar setting may arise in the context of crowdsourcing. 
For instance, professional fact-checkers are ideal for determining a gold standard label for the credibility of a news article \cite{allen2020scaling}, but their assessments may only be available for a small subset of data, whereas large volumes of crowdsourced, possibly biased, annotations can easily be collected. Even if a universal, unbiased gold standard may be elusive, a set of gold standard assessments that reflect organizational goals and values may be feasible and---as we will show in this work--very useful. For example, in the context of recruiting, a panel of senior recruiters with unconstrained time could do an in-depth assessment of a pool of candidates, yielding a gold standard that reflects the type of candidates that an organization would like to recruit. This gold standard could then be used to assess the decisions of recruiters who are operating under time pressure, as operating under such pressure is known to exacerbate biases \cite{stepanikova2012racial}. 
Table~\ref{tab:Examples} shows examples of individual human decisions and sources of gold standard assessments.


In this work, we tackle the problem of measuring human decision bias in settings where there is only a small set of gold standard decisions available. Specifically, we focus on bias measured as disparities in a given type of error of interest (e.g. false positives or false negatives) across different demographic groups, such as different genders or races. We propose a ML-based methodology that utilizes historical data of human decision makers whose decision bias we want to evaluate and a small, often disjoint, pool of instances with gold standard annotations. 
We evaluate our method, as well as alternative benchmarks, under various settings including different types of human bias 
and a wide range of data domains. Our method consistently outperforms or achieves comparable results relative to the alternatives.

\renewcommand{\arraystretch}{1.5} 
\begin{table}[H]
   \caption{Examples of Biased Decisions}
    \label{tab:Examples}
    \centering
    \small
    \begin{tabular}{p{0.15\linewidth} | p{0.45\linewidth} |p{0.30\linewidth}}
    
    \hline\hline
    \bf{Domain} & \bf{Human Decision ($Y'$)} &  \bf{Gold Standard Decision ($Y$)}
     \\ \hline\hline
    Medicine \ 
     & Physician's treatment decisions \cite{petersen2002impact}. & Expert panel assessments \cite{GORDON2009205}.    \\\hline 
    Crowd Sourcing & Crowdsourced annotations for fact-checking \cite{schulz2020we}. & Professional fact-checkers \cite{allen2020scaling}.   \\\hline
     
    \end{tabular}
\end{table}

\setlength{\parindent}{20pt}
The proposed method has broad implications across various key domains, including management and ML. Having access to improved measures of bias can make it easier to have a proactive approach to prevent problematic and consequential issues stemming from biased decisions. Research has shown that it is crucial for individuals to be aware of their biases before adopting new, desirable behaviors \cite{bandura2000social,chapman2013physicians,green2007implicit}. The proposed approach can enable such awareness. As such, our work adds to the emerging literature on AI for management \cite{jain2021editorial}, proposing a novel way in which AI can help advance organizational goals and support better decision-making. 

Furthermore, the consequences of biased human assessments may extend beyond the immediate consequences of a decision when observed human decisions are used as labels to train Artificial Intelligence (AI) and ML models. The field of information systems (IS) acknowledges the valuable impact of AI-based outcomes and their implications \cite{kordzadeh2022algorithmic,mikalef2018big,chen2012business,martin2019designing}. However, there is a paucity of IS research focusing on assessing data bias and its negative consequences across diverse critical domains \cite{aagerfalk2022artificial}. 
Human-generated assessments are a prevalent source of labels used to train AI and ML models, which are increasingly used to aid decision-making \cite{zhang2018fairness,brennan2009evaluating,mahoney2007method,angwin2022machine}. This creates a pathway through which human bias can be perpetuated into these models. Societal biases encoded in human assessments have been highlighted as an important source of algorithmic unfairness (e.g. \cite{machine_learning_bias_general, violago2018ai}). For example, it has been found that labels used to train AI recruiting tools showed bias against women (e.g. \cite{dastin2018amazon, kodiyan2019overview}). 
Research efforts aimed at mitigating algorithmic bias have focused on \emph{induction} bias---the bias that seeps in during training---, but generally assume that labels are accurate, resulting in a lack of methodoloigies to mitigate label bias \cite{li2022more}. 
The proposed method to measure bias in human decisions can serve as an important building block to mitigate the risk of encoding decison makers' bias in algorithm predictions.

We list our core contributions below:
\begin{enumerate}

\item We propose the problem of estimating human decision bias when gold-standard labels are limited and only accessible for a possibly disjoint set of data. 

\item 
We propose a ML-based solution for this problem, backed by theoretical guarantees. Our approach utilizes observed human decisions alongside the scarce and separate set of gold standard labels to assess bias in human decisions. We highlight the significance of a theoretical guarantee underlying our approach by conducting a comparison between our method and a variant that does not have such guarantee. 

\item We empirically assess our methodology and measure its effectiveness against alternative approaches across various settings, considering different types of human bias and different data domains. Our findings indicate that the proposed approach achieves state-of-the-art results.


\end{enumerate}

\section{Related Work}
\label{sec:rel}

In this section we review three bodies of work that are closely related to our research. First, we provide a review of the literature on biased human decision-making, as the primary goal of our work is to propose a methodology to assess such biases. We then review the literature on ML-based methods to evaluate different properties of human decisions; our research contributes to this body of work by proposing an ML-based method to assess human decision bias. Finally, we review the literature on algorithmic fairness, which relates to our work in two ways: (1) our method makes use of advances in algorithmic fairness that counter bias introduced during training of an algorithm, which enables our method to isolate bias contained in the human decisions, and (2) human decisions are increasingly used as labels to train ML algorithms, hence our method can help alleviate algorithmic bias stemming from this source, constituting a contribution to the algorithmic fairness literature. 


\subsection{Biased Human Decision-making}
An important focus of research in Information Systems (IS) is human decision-making \cite{goes2013editor}. A growing body of research has identified human biases in decision-making across several important domains, such as healthcare \cite{markowitz2022gender}, fundraising \cite{younkin2018colorblind}, ridesharing markets \cite{greenwood2020unbecoming}, and hiring \cite{wade2020social}. For instance, examinations of linguistic disparities in medical records have revealed underlying gender and ethnicity biases, impacting the quality of care provided to different demographic groups \cite{markowitz2022gender}. 
Similarly, African American founders have been shown to be significantly less likely to receive funding compared to their white counterparts, even when their project characteristics are identical~\cite{younkin2018colorblind}. In the ridesharing market, Caucasian male raters are more likely to penalize female drivers for poor service~\cite{greenwood2020unbecoming}. Lastly, research has shown that hiring managers are influenced by job-irrelevant political content, with applicants sharing similar political views receiving higher hireability ratings \cite{wade2020social}.

Both professional experts and general public's decisions can be affected by bias \cite{frank2019human,teodorescu2021failures}. Experts are susceptible to bias, as evidenced by gender and racial disparities observed in the emergent evaluation of potential acute coronary syndrome by expert physicians \cite{musey2017gender} and service providers \cite{musey2017gender}. Additionally, persistent disparities in clinical decision-making have also been documented for patients presenting with chest pain, particularly among African American men and women, as well as uninsured individuals, who receive fewer diagnostic tests \cite{pezzin2007disparities,chang2007gender,chapman2013physicians}. In the workplace,  supervisors have been found to assign lower job performance ratings to black employees compared to white employees, which negatively impacts their career satisfaction \cite{igbaria1992organizational}.

Biases may also shape decisions in everyday life and non-expert assessments. User ratings, for example, are frequently shaped by factors such as self-selection and social influence, leading to an overrepresentation of extreme opinions \cite{guodong2015vocal}. Furthermore, partisan biases can undermine the reliability of fact-checking, causing crowd workers to produce biased judgments when verifying the credibility of news articles \cite{saeed2022crowdsourced,chung2023ai}. Biased decisions, whether made by experts or the general public, have far-reaching consequences for both individuals and society as a whole \cite{richeson2016toward}. 

A body of research explores strategies to mitigate various biases in human decision-making, including cognitive and data biases \cite{paulus2024interplay}, self-selection biases \cite{hu2017self}, racial bias \cite{younkin2018colorblind,rhue2022you} and gender bias \cite{greenwood2020unbecoming}. For instance, evidence suggests that mindfulness helps decision-makers stay open to new information and avoid reinforcing initial biases, thereby mitigating the harmful effects of cognitive and data biases~\cite{paulus2024interplay}. Similarly, adjusting prices may counteract the effects of self-selection biases in online reviews, which distort the true representation of product quality and influence customers' purchasing decisions \cite{hu2017self}. Like our study, \cite{rhue2022you} and \cite{younkin2018colorblind} focus on societal biases such as race and gender. \cite{rhue2022you} argues that eliminating racial cues associated with the creator and the project helps reduce bias in crowdfunding decisions, while \cite{younkin2018colorblind} suggests that promoting successful minority founders or highlighting project quality through endorsements can mitigate these disparities. Unlike these studies, which investigate how biases influence decisions and explore interventions to alleviate the problem, our work offers a reliable solution for \emph{measuring} decision-makers' biases.

The causes of human decision bias have also been extensively explored across various fields, including IS, psychology, and healthcare. In IS, research has identified factors that contribute to biased decision-making during crises, such as limited access to data, poor data quality, infrastructure destruction, and a tendency for individuals to favor information that supports their pre-existing beliefs \cite{paulus2024interplay}. Research has also studied causes of societal biases, which are the biases that we are primarily concerned with in this work. Time pressure, cognitive load, and lack of information---common in emergency department settings--- have been shown to increase the likelihood of physicians relying on biased heuristics, potentially exacerbating racial disparities in healthcare delivery \cite{dehon2017systematic}. In contrast, standardized protocols and fear of legal consequences can help counteract physicians' racial biases \cite{dehon2017systematic}. Psychological theory has also contributed to our understanding of what triggers biases; for instance, research suggests that gender biases are reinforced when individuals are evaluated for roles traditionally associated with a specific gender (e.g., men in leadership positions)~\cite{kunda1996forming}. 

Our research contributes to the body of work on human decision biases from a design science perspective. While existing research has studied causes of decision bias, characterized its effects, and explored interventions to mitigate it, our work  offers a reliable and scalable solution for measuring biases in human decisions.

\subsection{ML-Based Human Decision Evaluation}

Advancements in ML have considered the use of expert and non-expert decisions and annotations as a source to train machine learning models. However, much of this research pertains to contexts and problems different from those we address in our study. A significant portion of prior work has focused on enhancing the quality of data labels gathered from multiple humans, including both crowd workers \cite{dai2013pomdp, dalvi2013aggregating,wang2017cost} and expert workers \cite{zhang2012estimating}. Notably, most studies in this domain rely on aggregating multiple human decisions to estimate gold standard labels \cite{dawid1979maximum, zhang2012estimating, dai2013pomdp, dalvi2013aggregating,sheng2017majority, wang2017cost}. However, we argue that biases can be shared among the individuals we seek to evaluate, as societal stereotypes, for example, may be pervasive among the general public. Consequently, any approach that relies on the wisdom of crowds and aggregates labels from multiple humans to estimate a gold standard may mislead the final assessment of decision bias. As a result, we consider gold standard labels obtained through panels of experts that are convened under circumstances that are primed to mitigate bias (e.g. reduced time pressure, improved deliberation processes, stronger expertise), rather than by aggregating the decisions of the evaluated individuals.


Nonetheless, advances in ML that tackle the fact that human-generated labels may be incorrect serve as an important benchmark for the proposed method. In particular, a stream of research has focused on learning from noisy labels by developing methods to evaluate their quality \cite{tanno2019learning,northcutt2021confident}. While the primary objective of these methods is to improve the quality of ML models trained on noisy labels, they can also serve to evaluate the quality of human-generated labels, and thus constitute an important benchmark for our work. 
Perhaps the most relevant work in the context of learning from and assessing noisy labels is Confident Learning (CL) \cite{northcutt2021confident}, due to its emphasis on assessing label quality and the fact that is a state-of-the-art method widely deployed across large businesses and organizations\footnote{https://cleanlab.ai/casestudies/}. Notably, its use of predicted probabilities generated by ML models to estimate label quality closely aligns with our methodology. Specifically, CL detects and characterizes labeling errors in datasets by employing principles to prune noisy data. 
However, unlike our work, CL assumes that errors are random conditioned on label class, which is a noise structure that does not encompass biases. It is worth noting that CL is originally proposed as a \emph{pruning} method, but the method prunes instances when it infers that the corresponding labels are incorrect, hence it can be easily adapted to measure errors and used as benchmark, which allows us to empirically demonstrate how our work compares to the state-of-the-art of ML research on learning from noisy labels. 

\subsection{Sources and Remedies of Algorithmic Bias}

The proposed method leverages the fact that ML models are prone to replicating bias contained in training labels \cite{Obermeyer2019dissect, bolukbasi2016man, baskerville2019digital, berente2021managing, gupta2022questioning}. However, using this principle to assess the degree of bias contained in the labels requires us to disentangle the bias introduced during the learning process from the bias coming from the human labels themselves. We do so by leveraging recent research in algorithmic fairness. Specifically, we implement a group fairness strategy designed to counter bias introduced during training by enforcing fairness with respect to the observed labels via a post-processing approach \cite{hardt2016equality}. 

Importantly, the proposed approach builds on recent advances in algorithmic fairness and also constitutes a contribution to the algorithmic fairness literature. Bias in human labels is a pathway through which societal bias can penetrate into the machine learning models trained on them \cite{Obermeyer2019dissect, bolukbasi2016man, baskerville2019digital, berente2021managing, gupta2022questioning}. However, most methodologies to mitigate algorithmic bias have not tackled this source of bias \cite{teodorescu2021failures}, and instead they focus on mitigating bias introduced at training time by using the available labels as a gold standard against which to measure bias of the algorithmic predictions \cite{de2022algorithmic}. Our work contributes to the algorithmic fairness literature by providing a means to leverage a disjoint, small set of gold standard labels to assess label bias prior to model induction.

\section{Problem Formulation}
\label{subsec:problem_formulation}

We consider a set of $K$ human decision-makers, such as crowd labelers or domain experts, $H= \{H^1,...,H^K \}$, whose decisions $Y'=\{Y'^1,...,Y'^K\}$ are encoded in historical observational data. For a given human decision-maker $H^k$ (e.g., a domain expert or a crowd-sourcing worker), the human assessment for an instance $i$, ${Y'}_i^k \in\{0,1\}$, and its corresponding feature vector, $X_i^k \sim \mathbb{P}(\mathcal{X})$, are available. As part of the feature vector, each instance $i$ has a sensitive attribute $A_i$ associated to it (e.g. gender, race, etc), which we denote separately given its relevance for our method. This feature is categorical and we denote it as $A\in\{a,\sim a\}$, where $a$ is a feature value with respect to which we wish to perform the bias assessment. For example, if $A$ corresponds to gender, $a=woman$ allows us to assess whether humans exhibit bias against (or in favor of) women. Thus, each human has an associated set of instances $S_{H^k}=\{X_i^k,A_i^k, {Y'}_i^k \}_{i=1}^{n^k}$, where $n^k$ is the number of instances labeled by human $H^k$. The instance sets evaluated by multiple humans may not overlap, but they should be sampled i.i.d. from the distribution. For instance, evaluations of physicians' diagnostic decisions do not require them to all evaluate the same patient, but do require the patients they evaluate to be sampled from the same distribution.

In addition, we consider that a gold standard decision, $Y$, is available for a small set of instances, $GS=\{X_l,A_l, Y_l \}_{l=1}^{m}$. As its name indicates, this constitutes the \emph{gold standard} against which one would like to assess the decisions of humans $H$. For example, when assessing physicians, the gold standard may be set by expert panels; when assessing crowd workers, the gold standard may be set by experts. We assume that the set of instances for which the gold standard is available, $GS$, may not overlap with any of the humans' own decision sets, $S =\{S_{H^k}\}_{k =1}^K$, and may be substantially fewer, yet are sampled i.i.d. from the same distribution. 

\begin{figure}[H]
\centering
\includegraphics[width=0.7\textwidth]{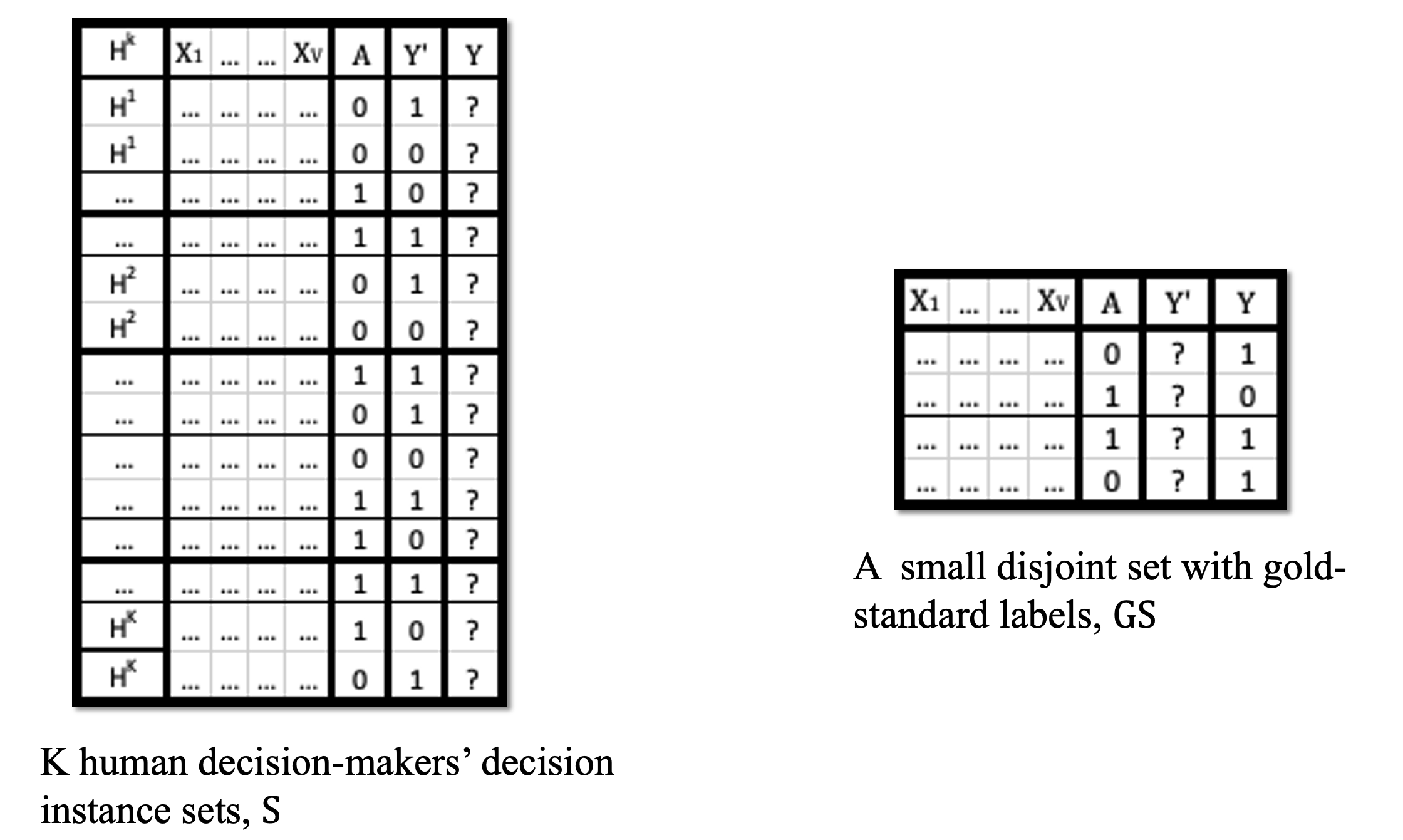}\caption{An illustration of K human decision-makers' decision instance sets (left), $S =\{S_{H^k}\}_{k =1}^K$  where  $S_{H^k}=\{X_i^k,A_i^k, {Y'}_i^k \}_{i=1}^{n^k}$ and a small non-overlapping set with gold-standard labels (right), $GS=\{X_l,A_l,Y_l \}_{l=1}^{m}$.}\label{fig:problem_setting}
\end{figure}

We seek to produce assessment of humans' decision biases. We define bias as disparities in rates of error for a certain type of error of interest (\cite{hardt2016equality}), such as gaps in true positive rates (TPRs), as shown in Equation Eq.\ref{eq:biasgap}, where $TPR\:_{Y'|Y, \:a}^k$ denotes the true positive rate of $Y'^k$ with respect to $Y$ for the subset of cases with sensitive attribute value $A=a$.   

\begin{equation}\label{eq:biasgap}
    GAP\:_{Y'|Y,\:A}^k = TPR\:_{Y'|Y, \:a}^k - TPR\:_{Y'|Y,\:\sim a}^k
\end{equation}

 We consider this measure throughout this paper, but analogous theoretical derivations and empirical results can be obtained when considering gaps in false positive rates (FPRs) across groups. 
Crucially, directly assessing disparities in error rates would require us to have a gold standard decision for every instance assessed by each human. The goal in this work is to leverage scarce and potentially disjoint gold standard data GS to assess bias in $S$. 
Table \ref{tab:notation} summarizes key notations used throughout the paper. 

\renewcommand{\arraystretch}{1.5} 
\begin{table}[htbp]
\singlespace
   \caption{Key Notations}
    \label{tab:notation}
    \centering
    \small
    \begin{tabular}{p{0.25\linewidth} |p{0.70\linewidth}}
    
    \hline\hline
    Notation & Description \\ \hline
     $X_i$ & The feature vector for an instance $i$ \\\hline
    $A_i$ & The sensitive attribute associated with an instance $i$ \\\hline
    $Y_i$ & Gold standard label of an instance $i$ \\\hline
    $GS=\{X_l,A_l, Y_l \}_{l=1}^{m}$ & A small set for which instances, sensitive attributes and gold standard labels are available \\\hline
    $H= \{H^1,...,H^K \}$ & Human decision-makers (e.g., crowd labelers or domain experts) of decisions to be evaluated \\\hline
     $Y'=\{Y'^1,...,Y'^K\}$ & Sets of human assessments of K humans, $H^1,...,H^k$, such that ${Y'}_j^k \in\{0,1\}$ corresponds to the decision of human $H^k$ for instance $j$\\\hline
      $S_{H^k=}\{X_i^k,A_i^k, {Y'}_i^k \}_{i=1}^{n^k}$ & Set of instances, sensitive attributes and decisions associated to human $H^k$\\\hline
       $S =\{S_{H^k}\}_{k =1}^K$ & Set of decision sets associated to humans $H$ \\\hline
    $f^k$  & A single base model, mapping $: X^k \rightarrow Y'^k$, which is trained on decisions by human $H^k$ \\\hline

    \end{tabular}
\end{table}

\section{Method}
\label{sec:method}

This section introduces the proposed methodology: \textbf{M}achine-learning-based human \textbf{D}ecision \textbf{B}ias \textbf{A}ssessment (MDBA). 
We first describe the method and provide intuition for it, and then provide theoretical reasoning and guarantees.

\subsection{\textbf{M}achine-learning-based human \textbf{D}ecision \textbf{B}ias \textbf{A}ssessment (MDBA)}
The proposed MDBA method first trains models to predict the decisions of each human decision maker, yielding a set of predictive models $\{f^k\}_{k=1}^{K}$, where each model is a mapping $f^k: X^k \mapsto Y'^k$, induced from the data set $S_{H^k}$ associated to human $H ^k$. In other words, there is one model to represent each human. It has been recognized that machine learning models absorb and reproduce biases contained in the data used to train them (\cite{barocas2016big, de2022algorithmic}). This is typically understood as a problem (\cite{barocas2016big,berente2021managing}), but in this work we leverage this property to our advantage. 
We ultimately aim to use the models $\{f^k\}_{k=1}^{K}$ to estimate the humans' decision biases. 

Biases contained in the models, however, will have multiple sources; in particular, some biases are introduced during model training and do not correspond to (but rather might compound) the humans' decision biases. Thus, the second step of the proposed methodology applies a bias mitigation strategy to counter bias introduced during the learning phase, which assesses disparate deviations of a model's prediction $\hat{Y}$ with respect to the human decision it is trained to predict, $Y'$. In recent years, a robust body of work has developed methodologies to mitigate bias introduced during training. Popular among them are post-processing strategies, which apply a transformation to the output of the algorithm or to the classification threshold to ensure a fairness constraint, such as equalizing false positive rates across groups (\cite{hardt2016equality}). We propose a recall-versus-precision ratio (RPR) constraint via post-processing. Specifically, we consider the group-specific recall (also known as true positive rate), and precision (also known as positive predictive value), 
of models' predictions $\hat{Y}$ with respect to the human labelers' decisions $Y'$ given a protected group, namely $TPR\:_{\hat{Y}|Y', \:A}$ and $PPV\:_{\hat{Y}|Y', \:A}$.
In the post-processing step, we recalibrate the classification thresholds for each model to equalize RPR across the set of models $\{f^k\}_{k=1}^{K}$. In cases where multiple thresholds yield the same RPR, i.e. there is not a unique mapping from RPR ratio to corresponding classification threshold, the algorithm can be ran iteratively using the multiple thresholds, and the bias can be estimated as the average of the different outputs. Naturally, high variance across the estimates is indicative of uncertainty. 

\begin{figure}[!htb]
\centering 
\caption{Key Steps of the MDBA methodt}\label{fig:method_key_ideas}
\includegraphics[width=1\textwidth]{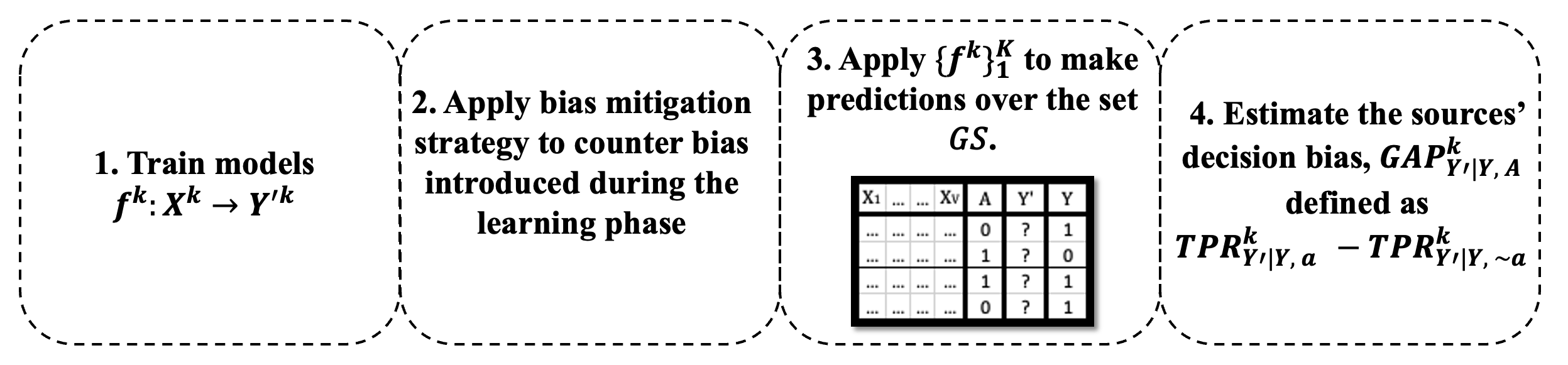}
\end{figure}

In a third step, the set of adjusted models, denoted as $\{f^k\}_{k=1}^{K}$, is used to make predictions on the disjoint and scarce gold standard dataset $GS$. Intuitively, this provides an estimate of the decision each human would have made on each instance in this set. Finally, in the fourth step we assess bias of $\{f^k\}_{k=1}^{K}$ with respect to $Y$, i.e., $GAP\:_{\hat{Y}|Y,\:A}$ as defined in Equation \ref{eq:biasgap}, which yields an estimate of the humans' decision bias. Figure \ref{fig:method_key_ideas} shows the four key steps in our approach, and the complete procedure is detailed in Algorithm~\ref{alg:mba_version3}.

\begin{algorithm2e}[!htb]
\caption{MDBA Algorithm}
\LinesNumbered
\SetAlFnt{\small\sf}
\label{alg:mba_version3}
\DontPrintSemicolon  

  \SetKwFunction{FMBA}{MDBA}
 \SetKwProg{Fn}{Algorithm}{:}{}
  \Fn{\FMBA{$\{S_{H^k}\}_{k=1}^K$, $GS$, $c$}}{
    \BlankLine
         \textbf{foreach} $S_{H^k} \in \{S_{H^k}\}_{k=1}^K$ \textbf{do} train base model $f^k$ on $S_{H^k}$ \tcp*[f]{Step 1 ends}\;
    \BlankLine
    $\{\pi^k_{A=a}, \pi^k_{A=\sim a}\}_{k=1}^K \gets$ thresholds of $\{f^k\}_{k=1}^K$ satisfying $c$ \tcp*[f]{Step 2 ends}\;
    \ForEach{$f^k \in \{f^k\}_{k=1}^K$}{
    $\{\hat{Y}\}_{l=1}^m \gets$ use $f^k$ with $[\pi^k_{A=a}, \pi^k_{A=\sim a}]$ classify $GS = \{X_l, Y_l\}_{l=1}^m$ \;
    $GAP\:_{\hat{Y}|Y,\:A}^k = TPR\:_{\hat{Y}|Y, \:a}^k - TPR\:_{\hat{Y}|Y,\:\sim a}^k$}
    \BlankLine
        \KwRet $\{GAP\:_{\hat{Y}|Y,\:A}\}_{k=1}^K$ \tcp*[f]{Step 3 and 4 end}
        }
\end{algorithm2e}

\subsection{Theoretical Analysis}
\label{subsec:theoretical_analysis}
We now show that, given the correct functional form specification of the humans' models, i.e. assuming that the functional form of the relationship between the dependent variables and the human decisions, $f: X \mapsto Y'$, is correctly specified, our method can accurately assess humans' decision bias. We demonstrate this in the series of theorems below. The proofs of the lemma and the theorems can be found in Appendix~\ref{section:Proof}.

\begin{lemma}
\label{lem:lemma1}
Given the correct functional form specification of each human's decision model $f^k: X \mapsto Y'^k$, then $\hat{Y} \independent Y | Y'^k$ and ${Y}'^k \independent Y | \hat{Y}$. 
\end{lemma}
\begin{theorem}
\label{theorem}
Given the correct functional form for the humans' decision models ($f_k: X \rightarrow Y'^k$), then there exists a ratio $\frac{TPR_{\:\hat{Y}|Y', \:A}^l}{PPV_{\:\hat{Y}|Y', \:A}^l} = \frac{TPR_{\:\hat{Y}|Y', \:A}^k}{PPV_{\:\hat{Y}|Y', \:A}^k} = c$, equivalent to $\tfrac{|\hat{Y}^l=1, A = a, \sim a|}{|Y'^l=1, A = a, \sim a|} = \tfrac{|\hat{Y}^k=1, A = a, \sim a|}{|Y'^k=1, A = a, \sim a|} = c$, such that if the biases exhibited by a pair of  models of human decisions, $f_l$ and $f_k$, are such that $GAP\:_{\hat{Y}|Y,\:A}^l >
GAP\:_{\hat{Y}|Y,\:A}^k$, then the decision biases of this pair of humans are such that $
GAP\:_{Y'|Y,\:A}^l > GAP\:_{Y'|Y,\:A}^k$. 
\end{theorem}

\begin{theorem}
\label{theorem2}
Given the correct functional form for the humans' decision models ($f: X \rightarrow Y'$), then there exits a ratio $\frac{TPR_{\:\hat{Y}|Y', \:A}}{PPV_{\:\hat{Y}|Y', \:A}} = c$ such that the bias in the human's model, $GAP\:_{\hat{Y}|Y,\:A}$, and decision bias of the human, $GAP\:_{Y'|Y,\:A}$, follow the relationship: $\frac{GAP\:_{\hat{Y}|Y,\:A}}{c} = GAP\:_{Y'|Y,\:A}$, where $
GAP\:_{\hat{Y}|Y,\:A} = TPR\:_{\hat{Y}|Y, \:a} - TPR\:_{\hat{Y}|Y,\:\sim a}$
and $GAP\:_{Y'|Y,\:A} = TPR\:_{Y'|Y, \:a} - TPR\:_{Y'|Y,\:\sim a}$.
\end{theorem}

The complete proof is presented in Appendix A. Theorem \ref{theorem} allow us to construct the relative decision biases of the humans. Theorem \ref{theorem2} derives the solution of humans' actual decision bias assessment. 


Theorem \ref{theorem} indicates that the value of c correspond to the ratio between the number of positive predictions made by a human's decision model $\hat{Y}=1$ for group $A = a$ or $A =\sim a$, and the actual positive decisions made by the human $Y'=1$ for the same groups. By adjusting the threshold of each model used to assign positive labels, multiple c values can be obtained. Each value of c corresponds to a different probability threshold for each model and protected group. 
According to Theorem \ref{theorem}, by adjusting the probability threshold of each model to achieve RPR ratio = c, it is possible to recover the ranking of decision biases by using the human models' prediction biases. Theorem \ref{theorem2} implies that further dividing the human models' prediction biases by the parameter c will achieve the human's actual decision bias assessment. Note that prior to enforcing the desired threshold on all the humans' decision models, each model has an initial threshold pair for each protected group variable value, given by $\{\pi'_{A=a}, \pi'_{A=\sim a}\} = \{0.5, 0.5\}$. The ultimate threshold pairs, given by $\{\pi_{A = a}^k, \pi_{A = \sim a}^k\}$ for the human $H^k$, are the threshold pairs that satisfy the PRP ratio = c. 

\section{Empirical Evaluation}
\label{sec:empiric}

 To evaluate our method, we conducted empirical evaluations using simulation studies across various publicly available datasets. Simulation studies are advantageous for creating a controlled computational environment that allows for the manipulation of experimental variables and the observation of outcomes  \cite{miranda2022editor,nan2017unifying}. Therefore, in line with established practices in related research  \cite{geva2021better,ipeirotis2014repeated,raykar2010learning,liebman2019right,geva2019using}, this paper compares the proposed method with alternative benchmarks by simulating a variety of conditions, including varying magnitudes of human decision biases, different class distributions, and various \emph{types} of biases. The proposed evaluation, for which we provide code\footnote{Anonymized for submission}, constitutes a publicly available evaluation framework that can be used by future approaches meant to assess human biases. In this section, we describe the empirical evaluation in detail. %

It is crucial to highlight that while our theoretical analysis offers guarantees when the functional form specification of the humans' decision models is accurate, our empirical evaluation does not rely on this assumption, and instead the simulations are purposefully constructed to \emph{violate} this assumption. The results demonstrate that even without knowledge of the correct functional form 
our approach remains effective.

\subsection{Datasets}

We consider four publicly available datasets for our experiments: (1) Adult, also known as ``Census Income" \footnote{https://www.kaggle.com/datasets/uciml/adult-census-income} dataset, in which the prediction task is to determine whether a person makes a high or a low income; (2) Credit dataset \footnote{https://archive-beta.ics.uci.edu/dataset/350/default+of+credit+card+clients}, in which the task is to predict default payments of credit card clients; (3) Employee Evaluation for Promotion\footnote{ https://www.kaggle.com/muhammadimran112233/employees-evaluation-for-promotion}, in which the task is to predict if a person is eligible for promotion or not; and (d) Hospital Readmission Rates dataset \footnote{https://www.kaggle.com/code/iabhishekofficial/prediction-on-hospital-readmission}, in which the task is to predict which patients will be readmitted. 

In all cases, we make use of the real covariates present in the data, thus ensuring that we are considering naturally occurring covariate distributions. The decisions and labels are simulated according to a variety of settings and conditions, which allows us to study the benefits and pitfalls of the different methods under different assumptions, as described below.

\subsection{Gold standard labels simulations}

In order to evaluate the performance of the different methods under different class distributions, we consider two scenarios: a positive label prevalence of 20\% and 30\%, respectively. We assume the prevalence of positive labels is constant across sensitive groups, a design choice that is meant to favor the widely used benchmark of comparing selection rates, since expecting unbiased decisions to yield equal selection rates implicitly assumes equal class prevalence across groups.   

We achieve the desired prevalence by first fitting a machine learning model using the original labels. We then sort the predicted probabilities in descending order and assign positive labels to the top 20\% instances of each group. The same process is applied to achieve a 30\% prevalence of positive labels. The two distributions we consider correspond to settings in which the positive class is smaller, which often arise in practice and pose additional challenges from a machine learning perspective. Examples of such imbalance are common: the proportion of patients who require certain treatment may be smaller than those who do not require it, and the proportion of those who are eligible for a job interview or a bank loan may be smaller than the proportion of those who are ineligible. 
Note that we purposefully use a different functional form specification to simulate the labels and to apply our method. While we simulate gold standard labels using Logistic Regression \cite{nick2007logistic}, we use Xgboost \cite{chen2016xgboost} to build models $f_k$ to predict human labels as part of the bias estimation. This means that we purposefully create all simulations such that the core assumption underlying the theoretical guarantees of our method is violated. This allows us to empirically validate that the proposed approach exhibits good performance even when assumptions are not met for its theoretical guarantees to hold. 

Finally, we assume the availability of a small, disjoint, pool of instances with gold standard labels, such as those provided by expert panels. To simulate this, we randomly select a pool of the gold-standard labels generated, and we make these available to both our method and to the alternatives to leverage for the decision bias assessment. In order to study how varying amounts of gold standard labels impact the performance of the methods, we perform multiple simulations in which we vary the size of this pool. To ensure that differences in performance are due to the size of this pool and are not affected by the amount of decisions available for each human, a total of 800 instances (400 from each group) stratified by class (i.e. class prevalence corresponding to that of the overall data) are randomly selected from the dataset, and constitute a separate set from which we sample the pool of gold-standard labels available for each experiment. We begin with a pool of 100 instances per group and then incrementally add 100 instances at a time. This results in different settings of 100, 200, 300, and 400 instances with synthetic gold-standard labels from each group.

\subsection{Decision Simulations}

We assume that instances are randomly allocated to human decision makers. We simulate this by performing a stratified random  partition of the dataset into $K$ subsets. Note that this excludes the set for which we assume that only gold-standard labels are available. We then simulate decisions for each human according to the procedures described below. 

In order to simulate human decisions, we first note that when bias is defined as group disparities in a certain type of error, any set of biased decisions can be categorized in one of two buckets: \emph{correct within-group ordering} and \emph{incorrect within-group ordering}. This categorization refers to the relative (implicit) ranking of instances. In the \emph{correct within-group ordering} category, given two instances that belong to the same group, the one that is most qualified according to a criteria relevant to the decision at hand is more likely to be selected. For example, if the task is to select candidates to interview for a job and we are concerned with gender bias against women, \emph{correct within-group ordering} is respected if women who are selected are more qualified for the job than those women who are not selected. In such setting, bias would arise if the ``bar" for men an women is different; in other words, if there are men who are selected who are less qualified than some of the women who were not selected. This type of bias may arise across domains; for instance, widespread erroneous beliefs about black people's nerve endings being less sensitive than white people's can result in biases in pain management \cite{hoffman2016racial}. 

In the case of \emph{incorrect within-group ordering}, given two candidates that belong to the same group, the less qualified among them may be more likely to be selected. In other words, the implicit ranking within each group is incorrect. For example, in the context of workplace bias, women can be perceived negatively when exhibiting certain leadership attributes \cite{heilman2012gender}. 
Similarly, in the context of healthcare diagnosis such biases can emerge when diagnostic criteria is driven by disease presentation in one group and has limited validity for others, such as dermatological conditions that present differently across different skin colors \cite{adamson2018machine}. 
In our experiments, we simulate scenarios corresponding to both types of biases. We provide an overview of both simulation procedures below, with the pseudocode for simulating biased human decisions for the sensitive group (e.g., $A = women$) included in Appendix \ref{sec:simulation_code}.



\paragraph{Correct within-group ordering.} We simulate $K = 10$ human decision makers with varying degrees of bias. Specifically, we vary the true positive rate of human decisions with respect to the gold standard labels within the women group ($TPR_{Y'|Y,\:A=women}$), while keeping $TPR_{Y'|Y,\:A=men}$ fixed. We consider a range from 0.54 to 0.90, with equal intervals between each adjacent error rate. In other words, the humans with the least and the most bias have $TPR_{Y'|Y,\:A=women}=0.9$ and $TPR_{Y'|Y,\:A=women}=0.54$, respectively. To achieve this, we train $K$ machine learning models that predict the gold standard label $Y$, each using the instance set associated to a human, $H^k$. We then obtain the simulated decisions $Y'$ associated to women by adjusting the standard classification threshold (0.5) to a desired threshold that yields the predefined $TPR_{Y'|Y,\:A=women} \pm 0.01$ for $S_{H^k}$. For the portion of instances associated to men, we assume that humans correctly assess them, except for random noise that yields a  $TPR_{Y'|Y,\:A=men}=0.95 \pm 0.01$.


\paragraph{Incorrect within-group ordering.} We again simulate $K=10$ human decision makers but this time we consider their misuse of an interaction term resulting in biased and incorrectly ordered decisions.  This comprises cases that yield an implicit ranking within the disadvantaged group that is incorrect and non-random. We simulate this type of bias by selecting a covariate, $Z$, which we assume is one that decision makers integrate incorrectly into their decisions; we then create a new covariate defined as the interaction $Z \times A$.\footnote{The interaction terms which we considered for different datasets include (1) Adult: $\textit{sex} \times \textit{education}$; (2) Credit: $\textit{sex} \times \textit{marital status}$; (3) Employee: $\textit{sex} \times \textit{previous year rating}$; (4) Readmission: $\textit{sex} \times \textit{admission type}$.} 
 We use the new set of covariates to train $K$ machine learning models that predict the gold standard label $Y$, each using the instance set associated to a human, $H^k$. We then iteratively manipulate the coefficient associated to the interaction term for each human, such that we meet a predefined $TPR_{Y'|Y,\:A=women}$. 
 We repeat this process to simulate biased decision sets for all human decision makers, such that the resulting values for $TPR_{Y'|Y,\:A=women}$ range from 0.5 to 0.9. For the portion of instances associated to men, we assume that humans correctly assess them, except for random noise that yields a  $TPR_{Y'|Y,\:A=men}=0.95 \pm 0.01$.

\subsection{Benchmarks}

We evaluated our proposed approach relative to a set of benchmarks that represents diverse possibilities to tackle the problem of bias assessment. While the problem of estimating decision bias by leveraging a small and non-overlapping set of gold standard labels has not been tackled in the past, there are different ways in which one can make use of existing methodologies from different streams of literature to yield a bias estimate. Below we explain each one of the benchmarks considered.

\paragraph{Selection Rates (SR).}SR is perhaps the most intuitive and widely considered measure in practice  \cite{mehrabi2021survey} when gold standard labels are unavailable. Specifically,  SR estimates a human's bias by measuring the difference between the proportion of positive labels that the human assigns to instances from different groups. For example, the difference of promotion rates among men and women employees. The SR gap is formally defined in Equation~\ref{eq:sr}.
\begin{equation}
\label{eq:sr}
    \widehat{GAP}_{sr}^k = \sum_{i=1}^{|S_{H^k}|} I[Y'^k=1|A = a] - \sum_{i=1}^{|S_{H^k}|} I[Y'^k=1|A = \sim a]
\end{equation}

\paragraph{Gold Standard (GS) -based.} This benchmark, refereed as ``GS-based'', uses a predictive model\footnote{We use a well-known methodology---fine-tuned XGBoost  \cite{chen2016xgboost}---as the predictive modeling technique in both GS-based and CL to infer the correct labels. This enables the most direct head-to-head comparison with our method, as this is the same base model we use to implement the proposed approach in the experiments.} trained to predict the gold standard label on the scarce set of instances for which this label is available. A GS-based estimate of humans' decision bias, referred as $\widehat{GAP}_{gs-based}^k$, is estimated through a comparison between the human decisions $Y'$ and those inferred by the predictive model, $\hat{Y}_{gs-based}$. 
\begin{equation}\label{eq:gsBased}
    \widehat{GAP}_{gs-based}^k = TPR\:_{Y'|\hat{Y}_{gs-based}, \:a}^k - TPR\:_{Y'|\hat{Y}_{gs-based}, \:\sim a}^k
\end{equation}
\paragraph{Confident Learning (CL).} Confident Learning (CL) represents the state of the art in a stream of literature that seeks to train machine learning models using noisy labels \cite{northcutt2021confident}. While this stream of research has not explicitly considered the type of bias we are concerned with, nor has it been proposed as a means to assess decision bias, it is a reasonable baseline insofar as it allows us to assess the performance of existing methods that assume labels are imperfect when repurposed for our goal. CL uses noisy labels to estimate the joint distribution between noisy (given) labels and gold standard (unknown) labels. The method then leverages the estimated probability distributions to prune the noisy data, producing clean data as output. To apply CL to our problem, we merge all labels available, encompassing both the disjoint pool of gold standard data and the humans' noisy decisions, and then apply CL across the complete dataset. 
We then compare the humans' decisions $Y'$, with the predicted labels $\hat{Y}_{CL}$ to assess the bias in their decision-making, refereed as $\widehat{GAP}_{CL}^k$.
\begin{equation}\label{eq:cl}
    \widehat{GAP}_{CL}^k = TPR\:_{Y'|\hat{Y}_{CL}, \:a}^k - TPR\:_{Y'|\hat{Y}_{CL}, \:\sim a}^k
\end{equation}

\section{Result}
\label{sec:result}

In this section, we assess the performance of the proposed MDBA approach  and compare it with that of the  benchmarks, SR, GS-based and CL. Following the evaluation framework introduced in Section~\ref{sec:empiric}, we consider different types of biases, different positive label prevalence (class distributions), various data domains, and varying amounts of gold standard instances available in the disjoint pool.  

For each setting, we apply all methods and evaluate them by computing the mean absolute error (MAE) between the assessment of a given approach and the true bias of the humans. In order to compute confidence bounds, we perform 20 iterations of each experiment, and report average MAE along with their 90\% confidence bounds. We then numerically report the relative improvements of the proposed approach relative to the alternatives, and assess its statistical significance.

\begin{figure}[!htb]
\centering
\caption{MDBA's performance relative to benchmarks under scenarios where 
humans exhibit {\itshape correct} within-group orderings and datasets are with 20\% and 30\% positive label prevalence.}
\BlankLine
\includegraphics[width=\textwidth]{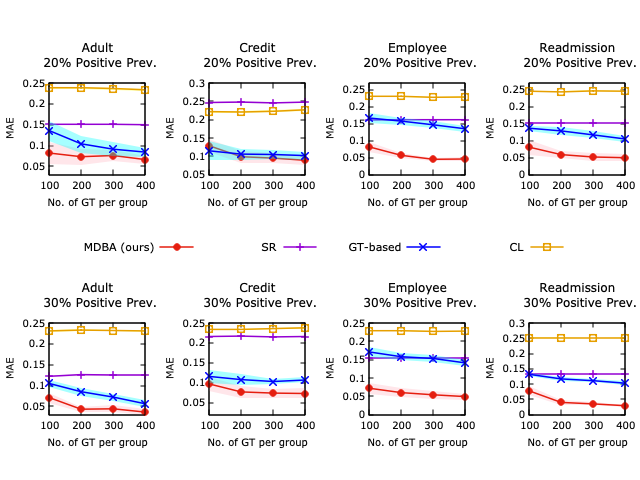}

 \begin{flushleft}
  \small
\subcaption*{MAE measure for humans' decision bias estimation errors for our approach MDBA and the baseline approaches. Results are reported given a varying number of gold standard (GS) instances, different datasets with different positive label prevalence. The shaded region shows the 95\% confidence bound for each method.} 
\end{flushleft}
\label{fig:main_correctWithinGroup}
\end{figure}

We first evaluate the methods for the \emph{correct within-group ordering} scenarios, in which the relative ranking is correct within each group, but the decision threshold is miscalibrated across groups. The detailed description of these scenarios can be found in Section~\ref{sec:empiric}. Figure \ref{fig:main_correctWithinGroup}, Tables \ref{tab:easyforsr_2020perc} and \ref{tab:easyforsr_3030perc} display the MAE measures attained by our method, MDBA, alongside the three benchmarks. 
Tables \ref{tab:easyforsr_2020perc} and \ref{tab:easyforsr_3030perc} also present the percentage improvement achieved by MDBA compared to each benchmark and its statistical significance. 
These results illustrate the distinct superiority of our method, which consistently outperforms the alternatives. There is only one dataset (Credit) at a specific label prevalence (20\%) in which one other method (GS-based) yields comparable performance to the proposed MDBA. Across all other scenarios, the proposed method yields statistically significant improvements.   

In Figure \ref{fig:main_correctWithinGroup}, the curves of MAE measures, along with their 90\% confidence bounds, show the performance of each approach as the number of ground truth labels available varies. SR does not make use of the gold standard labels, so it is expected that its performance does not change depending on the size of the gold standard set. GS-based, CL and MDBA do use these labels, but their sensitivity to the magnitude of this set differs. For CL, these results suggest that the signal contained in this small set is overpowered by the data coming from human decisions in its estimate of the distribution of a latent gold standard, yielding stable but relatively poor performance across all settings. Meanwhile GS-based and MDBA both exhibit improved performance as the amount of gold standard labels improve. Notably, while the trend of both methods is similar, MDBA yields consistently better performance. As shown in Tables~\ref{tab:easyforsr_2020perc} and \ref{tab:easyforsr_3030perc}, the improvement of MDBA relative to GS-based is frequently over 50\%, suggesting that MDBA makes a more effective use of the information available. Moreover, the improvement in performance as a function of relatively small changes in the number of gold standard labels available is a valuable property for organizations, as it indicates that small investments to obtain gold standard labels (e.g. an expert panel that labels 100 additional instances) can yield a high return in terms of improved performance.



\begin{table}[htbp]
\singlespacing
\begin{center}
\pgfplotstableset{highlight/.append style={
    postproc cell content/.append code={
            \pgfkeysalso{@cell content=\textbf{##1}}%
    }
}}

\begin{threeparttable}
\singlespacing
\caption{MDBA and benchmarks performance measured by MAE for humans exhibiting {\itshape correct} within-group orderings and datasets are with 20\% positive label prevalence}
    \label{tab:easyforsr_2020perc}
    \sisetup{
        round-mode          = places, 
        round-precision     = 2, 
    }
    \begin{filecontents*}{EasyForSR_2020perc_CLcorrected.csv}
Dataset,GT Per Group,MDBA,SR,MDBA_improv,GT-based,MDBA_improv,CL,MDBA_improv
Adult,100,0.082,0.151,45.4\%***,0.136,39.5\%***,0.239,65.5\%***
,200,0.073,0.151,51.8\%***,0.104,29.7\%**,0.239,69.5\%***
,300,0.076,0.151,49.3\%***,0.091,16.4\%,0.237,67.7\%***
,400,0.066,0.15,55.9\%***,0.084,21.3\%**,0.234,71.8\%***
Credit,100,\textnormal{0.129},0.247,47.5\%***,\textbf{0.115},-12.1\%,0.222,41.6\%***
,200,0.099,0.248,60.2\%***,0.106,6.7\%,0.221,55.4\%***
,300,0.095,0.246,61.3\%***,0.105,9.3\%,0.223,57.3\%***
,400,0.089,0.248,63.9\%***,0.102,12\%,0.227,60.6\%***
Employee,100,0.082,0.161,49.1\%***,0.167,51\%***,0.231,64.5\%***
,200,0.057,0.162,64.8\%***,0.158,63.9\%***,0.231,75.4\%***
,300,0.045,0.162,72.4\%***,0.147,69.5\%***,0.228,80.3\%***
,400,0.046,0.162,71.5\%***,0.135,65.6\%***,0.229,79.8\%***
Readmission,100,0.081,0.152,46.6\%***,0.137,40.9\%***,0.246,67.1\%***
,200,0.059,0.152,61.4\%***,0.129,54.7\%***,0.244,76.0\%***
,300,0.052,0.152,66.2\%***,0.117,56\%***,0.247,79.2\%***
,400,0.05,0.152,67.4\%***,0.105,52.8\%***,0.246,79.9\%***
\end{filecontents*}
\pgfplotstabletypeset[
    font=\normalsize,
      multicolumn names, 
      col sep=comma, 
      column type=l,
      display columns/0/.style={string type ={ S},column name={Dataset}, column type=c|},  
      display columns/1/.style={string type ={S},column name={\textsc{GS Per}}, column type=c|},
      display columns/2/.style={highlight, string type ={S},column name={MDBA}, column type=c|},
      display columns/3/.style={string type ={S},column name={SR}},
      display columns/4/.style={string type ={S},column name={MDBA}, column type=c|},
      display columns/5/.style={string type ={S},column name={GS-}},
      display columns/6/.style={string type ={S},column name={MDBA}, column type=c|},
      display columns/7/.style={string type ={S},column name={CL}},
      display columns/8/.style={string type ={S},column name={MDBA}, column type=c|},
      every head row/.style={
		before row={\toprule}, 
		after row={& Worker & (ours) &  &IMPROV. &based &IMPROV. &  & IMPROV.\\
			\midrule } 
			},
		every last row/.style={after row=\bottomrule}, 
            every row no 4/.style={before row=\midrule},
		every row no 8/.style={before row=\midrule},
            every row no 12/.style={before row=\midrule},
		every row no 16/.style={before row=\midrule}
    ]{EasyForSR_2020perc_CLcorrected.csv} 

  \begin{tablenotes}\small
  \linespread{1}
          \item\hspace*{-\fontdimen2\font}humans' bias estimation errors. Values show Mean Absolute Error (MAE). MDBA IMPROV. shows the improvement of MDBA over the alternatives.  MDBA yields substantially better and otherwise comparable estimations of workers accuracies. *** statistically significantly better ($p<0.01$), **: ($p<0.05$), and *: ($p<0.1$). The negative MDBA IMPROV (-12.1\%) indicates that GS-based is comparable to MDBA, with no statistically significant difference found between them.
    \end{tablenotes}
        
    \end{threeparttable}
    \end{center}
\end{table}

\begin{table}[htbp]
\begin{center}
\pgfplotstableset{highlight/.append style={
    postproc cell content/.append code={
            \pgfkeysalso{@cell content=\textbf{##1}}%
    }
}}

\begin{threeparttable}
\singlespacing
\caption{MDBA and benchmarks performance measured by MAE for humans exhibiting {\itshape correct} within-group orderings and datasets are with 30\% positive label prevalence}
    \label{tab:easyforsr_3030perc}
    \sisetup{
        round-mode          = places, 
        round-precision     = 2, 
    }
    \begin{filecontents*}{EasyForSR_3030perc_CLcorrected.csv}
Dataset,GT Per Group,MDBA,SR,MDBA_improv,GT-based,MDBA_improv,CL,MDBA_improv
Adult,100,0.071,0.122,42.1\%***,0.105,32.4\%***,0.231,69.3\%***
,200,0.043,0.126,66.2\%***,0.085,49.6\%***,0.233,81.7\%***
,300,0.044,0.125,65.1\%***,0.072,39.2\%***,0.232,81.2\%***
,400,0.036,0.125,70.8\%***,0.056,35.2\%***,0.231,84.2\%***
Credit,100,0.097,0.216,54.9\%***,0.116,16.1\%*,0.234,58.5\%***
,200,0.077,0.217,64.5\%***,0.108,28.4\%***,0.234,67.0\%***
,300,0.074,0.215,65.6\%***,0.103,28.2\%***,0.236,68.6\%***
,400,0.073,0.216,66.2\%***,0.107,31.8\%***,0.238,69.2\%***
Employee,100,0.072,0.155,53.5\%***,0.171,57.7\%***,0.229,68.5\%***
,200,0.06,0.155,61.5\%***,0.158,62.3\%***,0.229,73.9\%***
,300,0.054,0.155,65.6\%***,0.153,65.1\%***,0.227,76.4\%***
,400,0.049,0.155,68.3\%***,0.141,65.2\%***,0.228,78.5\%***
Readmission,100,0.078,0.133,41.3\%***,0.132,40.7\%***,0.252,69.0\%***
,200,0.04,0.133,69.9\%***,0.117,65.9\%***,0.252,84.2\%***
,300,0.035,0.133,73.7\%***,0.111,68.4\%***,0.252,86.1\%***
,400,0.029,0.133,78.1\%***,0.103,71.9\%***,0.252,88.5\%***
\end{filecontents*}
\pgfplotstabletypeset[
    font=\normalsize,
      multicolumn names, 
      col sep=comma, 
      column type=l,
      display columns/0/.style={string type ={ S},column name={Dataset}, column type=c|},  
      display columns/1/.style={string type ={S},column name={\textsc{GS Per}}, column type=c|},
      display columns/2/.style={highlight, string type ={S},column name={MDBA}, column type=c|},
      display columns/3/.style={string type ={S},column name={SR}},
      display columns/4/.style={string type ={S},column name={MDBA}, column type=c|},
      display columns/5/.style={string type ={S},column name={GS-}},
      display columns/6/.style={string type ={S},column name={MDBA}, column type=c|},
      display columns/7/.style={string type ={S},column name={CL}},
      display columns/8/.style={string type ={S},column name={MDBA}, column type=c|},
      every head row/.style={
		before row={\toprule}, 
		after row={& Worker & (ours) &  &IMPROV. &based &IMPROV. &  & IMPROV.\\
			\midrule } 
			},
		every last row/.style={after row=\bottomrule}, 
            every row no 4/.style={before row=\midrule},
		every row no 8/.style={before row=\midrule},
            every row no 12/.style={before row=\midrule},
		every row no 16/.style={before row=\midrule}
    ]{EasyForSR_3030perc_CLcorrected.csv} 

  \begin{tablenotes}\small
  \linespread{1}
          \item\hspace*{-\fontdimen2\font}humans' bias estimation errors. Values show Mean Absolute Error (MAE). MDBA IMPROV. shows the improvement of MDBA over the alternatives.  MDBA yields substantially better and otherwise comparable estimations of workers accuracies. *** MDBA is statistically significantly better ($p<0.01$), **: ($p<0.05$), and *: ($p<0.1$).
    \end{tablenotes}
        
    \end{threeparttable}
    \end{center}
\end{table}

\begin{figure}[htbp]
\centering
\caption{MDBA's performance relative to benchmarks under scenarios where 
humans exhibit {\itshape incorrect} within-group orderings and datasets are with 20\% and 30\% positive label prevalence.}
\BlankLine
\includegraphics[width=\textwidth]{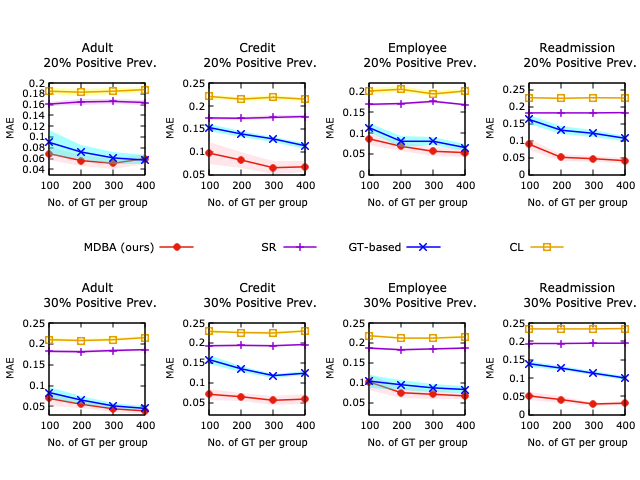}
 \begin{flushleft}
  \small
\subcaption*{MAE measure for humans' decision bias estimation errors for our approach MDBA and the baseline approaches. Results are reported given a varying number of gold standard (GS) instances, different datasets with different positive label prevalence. The shaded region shows the 95\% confidence bound for each method.} 
\end{flushleft}
\label{fig:main_IncorrectWithinGroup}
\end{figure}

Figure \ref{fig:main_IncorrectWithinGroup} and Table \ref{hardforsr_2020perc} and \ref{hardforsr_3030perc} show the results under the more challenging type of bias: \emph{incorrect within-group ordering}. Under this type of bias, humans' perceived ranking and calibration of decision thresholds both deviate from gold standard ranking and thresholds. Across all settings, the proposed method yields better or comparable performance to all baselines.

The percentual improvements achieved by the proposed method reach as high as 86.7\%, which we see in the case of the Readmission dataset, with a 30\% positive label prevalence and 300 instances in the disjoint gold standard set, when comparing with the CL baseline. In most cases, percentual improvements are over 50\%, which constitutes a drastic gain. When we consider the Adult dataset, in which we observe consistently comparable performance with the GS-based alternative, the magnitude of the MAE suggests that this is due to the structure of the data being sufficiently simple to learn a high-quality GS-based model, thus there are no significant gains from our proposed variant in this particular case.  

\begin{table}[htbp]
\begin{center}
\pgfplotstableset{highlight/.append style={
    postproc cell content/.append code={
            \pgfkeysalso{@cell content=\textbf{##1}}%
    }
}}

\begin{threeparttable}
\caption{MDBA and benchmarks performance measured by MAE for humans exhibiting {\itshape incorrect} within-group orderings and datasets are with 20\% positive label prevalence}
    \label{hardforsr_2020perc}
    \sisetup{
        round-mode          = places, 
        round-precision     = 2, 
    }
    \begin{filecontents*}{HardForSR_2020perc_CLcorrected.csv}
Dataset,GT Per Group,MDBA,SR,MDBA_improv,GT-based,MDBA_improv,CL,MDBA_improv
Adult,100,0.069,0.161,57\%***,0.09,23.2\%*,0.184988393,62.6\%***
,200,0.056,0.165,66\%***,0.072,22.2\%*,0.18300032,69.4\%***
,300,0.051,0.166,69.3\%***,0.061,17.1\%,0.18525603,72.5\%***
,400,\textnormal{0.06},0.164,63.7\%***,\textbf{0.057},-4\%,0.187624763,68.3\%***
Credit,100,0.097,0.174,43.9\%***,0.153,36.4\%***,0.220589558,55.9\%***
,200,0.082,0.173,52.7\%***,0.139,41.2\%***,0.215374237,62.0\%***
,300,0.065,0.175,62.8\%***,0.128,49\%***,0.218953179,70.3\%***
,400,0.067,0.177,62\%***,0.113,40.7\%***,0.215053099,68.7\%***
Employee,100,0.086,0.169,49.1\%***,0.113,23.6\%***,0.200871459,57.2\%***
,200,0.068,0.171,60.3\%***,0.08,15\%,0.205095498,66.9\%***
,300,0.056,0.176,68.4\%***,0.08,30.8\%***,0.19437147,71.4\%***
,400,0.053,0.168,68.6\%***,0.065,18.8\%*,0.201234597,73.7\%***
Readmission,100,0.09,0.182,50.7\%***,0.163,44.9\%***,0.226937724,60.4\%***
,200,0.051,0.182,71.9\%***,0.131,60.9\%***,0.225352509,77.3\%***
,300,0.047,0.182,74.1\%***,0.122,61.4\%***,0.226926704,79.2\%***
,400,0.041,0.183,77.7\%***,0.107,61.8\%***,0.226460118,82\%***
\end{filecontents*}
\pgfplotstabletypeset[
    font=\normalsize,
      multicolumn names, 
      col sep=comma, 
      column type=l,
      display columns/0/.style={string type ={ S},column name={Dataset}, column type=c|},  
      display columns/1/.style={string type ={S},column name={\textsc{GS Per}}, column type=c|},
      display columns/2/.style={highlight, string type ={S},column name={MDBA}, column type=c|},
      display columns/3/.style={string type ={S},column name={SR}},
      display columns/4/.style={string type ={S},column name={MDBA}, column type=c|},
      display columns/5/.style={string type ={S},column name={GS-}},
      display columns/6/.style={string type ={S},column name={MDBA}, column type=c|},
      display columns/7/.style={string type ={S},column name={CL}},
      display columns/8/.style={string type ={S},column name={MDBA}, column type=c|},
      every head row/.style={
		before row={\toprule}, 
		after row={& Worker & (ours) &  &IMPROV. &based &IMPROV. &  & IMPROV.\\
			\midrule } 
			},
		every last row/.style={after row=\bottomrule}, 
            every row no 4/.style={before row=\midrule},
		every row no 8/.style={before row=\midrule},
            every row no 12/.style={before row=\midrule},
		every row no 16/.style={before row=\midrule}
    ]{HardForSR_2020perc_CLcorrected.csv} 

  \begin{tablenotes}\small
  \linespread{1}
          \item\hspace*{-\fontdimen2\font}humans' bias estimation errors. Values show Mean Absolute Error (MAE). MDBA IMPROV. shows the improvement of MDBA over the alternatives.  MDBA yields substantially better and otherwise comparable estimations of workers accuracies. *** MDBA is statistically significantly better ($p<0.01$), **: ($p<0.05$), and *: ($p<0.1$). The negative MDBA IMPROV (-4\%) indicates that GS-based is comparable to MDBA, with no statistically significant difference found between them.
    \end{tablenotes}
        
    \end{threeparttable}
    \end{center}
\end{table}

\begin{table}[htbp]
\begin{center}
\pgfplotstableset{highlight/.append style={
    postproc cell content/.append code={
            \pgfkeysalso{@cell content=\textbf{##1}}%
    }
}}

\begin{threeparttable}
\singlespacing
\caption{MDBA and benchmarks performance measured by MAE for humans exhibiting {\itshape incorrect} within-group orderings and datasets are with 30\% positive label prevalence}
    \label{hardforsr_3030perc}
    \sisetup{
        round-mode          = places, 
        round-precision     = 2, 
    }
    \begin{filecontents*}{HardForSR_3030perc_CLcorrected.csv}
Dataset,GT Per Group,MDBA,SR,MDBA_improv,GT-based,MDBA_improv,CL,MDBA_improv
Adult,100,0.069,0.182,62.2\%***,0.083,16.6\%,0.210,67.2\%***
,200,0.055,0.181,69.8\%***,0.065,16.5\%*,0.208,73.7\%***
,300,0.044,0.184,76.1\%***,0.051,14.7\%*,0.210,79.2\%***
,400,0.039,0.186,79\%***,0.045,14\%,0.215,81.8\%***
Credit,100,0.071,0.193,63.3\%***,0.158,55.2\%***,0.229,69.0\%***
,200,0.065,0.194,66.6\%***,0.135,52.2\%***,0.226,71.4\%***
,300,0.056,0.193,71\%***,0.118,52.5\%***,0.225,75.2\%***
,400,0.059,0.196,69.9\%***,0.124,52.6\%***,0.230,74.4\%***
Employee,100,0.102,0.187,45.4\%***,0.104,1.7\%,0.218,53.0\%***
,200,0.075,0.183,58.9\%***,0.095,20.5\%**,0.212,64.5\%***
,300,0.071,0.185,61.8\%***,0.087,18.7\%**,0.212,66.7\%***
,400,0.067,0.187,64.4\%***,0.083,19.9\%***,0.215,69\%***
Readmission,100,0.051,0.194,73.7\%***,0.139,63.3\%***,0.234,78.2\%***
,200,0.041,0.194,79.1\%***,0.127,68.1\%***,0.234,82.7\%***
,300,0.029,0.195,85.3\%***,0.113,74.7\%***,0.234,87.8\%***
,400,0.031,0.195,84\%***,0.1,68.8\%***,0.235,86.7\%***
\end{filecontents*}
\pgfplotstabletypeset[
    font=\normalsize,
      multicolumn names, 
      col sep=comma, 
      column type=l,
      display columns/0/.style={string type ={ S},column name={Dataset}, column type=c|},  
      display columns/1/.style={string type ={S},column name={\textsc{GS Per}}, column type=c|},
      display columns/2/.style={highlight, string type ={S},column name={MDBA}, column type=c|},
      display columns/3/.style={string type ={S},column name={SR-}},
      display columns/4/.style={string type ={S},column name={MDBA}, column type=c|},
      display columns/5/.style={string type ={S},column name={GS-}},
      display columns/6/.style={string type ={S},column name={MDBA}, column type=c|},
      display columns/7/.style={string type ={S},column name={CL}},
      display columns/8/.style={string type ={S},column name={MDBA}, column type=c|},
      every head row/.style={
		before row={\toprule}, 
		after row={& Worker & (ours) &   & IMPROV. & based & IMPROV. &  & IMPROV.\\
			\midrule } 
			},
		every last row/.style={after row=\bottomrule}, 
            every row no 4/.style={before row=\midrule},
		every row no 8/.style={before row=\midrule},
            every row no 12/.style={before row=\midrule},
		every row no 16/.style={before row=\midrule}
    ]{HardForSR_3030perc_CLcorrected.csv} 

  \begin{tablenotes}\small
  \linespread{1}
          \item\hspace*{-\fontdimen2\font}humans' bias estimation errors. Values show Mean Absolute Error (MAE). MDBA IMPROV. shows the improvement of MDBA over the alternatives.  MDBA yields substantially better and otherwise comparable estimations of workers accuracies. *** MDBA is statistically significantly better ($p<0.01$), **: ($p<0.05$), and *: ($p<0.1$).
    \end{tablenotes}
        
    \end{threeparttable}
    \end{center}
\end{table}

It is noteworthy to highlight that the proposed method, MDBA, always outperforms what is arguably the most widely-used alternative, SR. We observe this to be the case for both correct and incorrect within-group ordering scenarios, with improvements ranging from 33\% to 83\%. The reason why SR fails is that it only considers the rate of selection across groups, ignoring which specific instances are truly qualified for the selection. Compared to the GS-based approach, our method excels in most settings and performs comparably in others. One of the key differences is that our method better leverages the historical decisions made by the humans to learn the patterns of their decision-making. In contrast, the goal of the GS-based approach is to predict the gold standard labels, which are scarce and therefore rarely sufficient to learn a good-enough model. Furthermore, MDBA applies bias mitigation techniques to address the algorithmic bias that may be introduced at training time, while the GS-based approach does not incorporate such bias mitigation with respect to the gold standard labels. Lastly, our method achieves an improvement ranging from 46\% to 88.5\% compared to the third benchmark CL, a state-of-the-art approach for learning from noisy labels. This approach assumes that noise is random conditioned on class. While accounting for class-conditioned noise is a big step forward in learning from noisy labels, and has been shown to yield very strong performance in domains such as image recognition (\cite{northcutt2021confident}), our results show that this approach is not suitable to assess bias in human labels. 
Overall, the results demonstrate the reliability and superior performance of our method.

\subsection{Ablation Study}

The method described in Section \ref{sec:method} shows that MDBA relies on a bias mitigation strategy to counter biases introduced during the learning phase. In this section, we present our ablation study to evaluate the relative benefits of this key element of our approach, specifically Step 2 in Figure \ref{fig:method_key_ideas}. We introduce a variant of our approach that follows Algorithms \ref{alg:mba_version3}, but without adjusting the thresholds for satisfying c values (line 3 in Algorithm 1). In this variant, when applying the models ($\{f^k\}_{k=1}^{K}$) to make predictions over the set GS, we use the initial thresholds for both groups, i.e., $\{\pi'_{A=a}, \pi'_{A=\sim a}\} = \{0.5, 0.5\}$. This variant is referred to as MDBA-Naive.

The results of the comparison between MDBA and MDBA-Naive are provided in Appendix \ref{sec:ablation_study}. Overall, MDBA consistently performs better or is comparable to MDBA-Naive, highlighting that Step 2, the bias mitigation strategy, is crucial in most cases and, at the very least, not detrimental.


\section{Discussion and Future Work}

In this paper, we tackle the problem of assessing biases encoded in human decisions. We propose a ML-based framework that returns an assessment of human decision biases, without requiring gold standard labels to be available for the instances assessed by the humans, nor any overlap in the instances assessed by different humans. The proposed approach estimates biases defined as gaps in true positive rates, and it can be extended to other bias metrics, such as gaps in true negative rates, among others. Assessment of such biases, which focus on disparities in \emph{quality} of the decisions across groups, are essential to capture biases that yield suboptimal outcomes for organizations and unfair treatment of individuals. We illustrate the performance of our approach by comparing it to several alternatives, including SR, which is widely used in practice, as well as two other machine learning-based methods: GS-based and CL. 
Our proposed approach achieves state-of-the-art results by effectively leveraging limited gold standard labels and incorporating humans' past decisions for which gold standard labels were missing. Crucially, the proposed method relies on the quality of models predicting expert labels (rather than gold standard labels); these models can be highly refined through tuning with the large volume of available human historical labels, and the quality of these models can be empirically validated. 
Our method accurately estimates human decision biases, outperforming all other alternatives, and thereby provides a solid foundation for reliable bias assessment in human decision-making.


Increasing transparency of human biases can offer a variety of benefits. Without the ability to measure biases, it is difficult to determine whether interventions aimed at curbing bias successfully improve fairness in decision-making tasks. The application of the proposed method would enable the assessment of decision biases both before and after such policies are implemented. This opens up opportunities for evidence-based interventions, and also open up interesting opportunities for research focused on designing policies that can truly make a difference. 

As noted in our work, there are contexts in which small sets of gold standard decisions are already available---such as the case of expert panels in healthcare---and the proposed method enables a way of productively utilizing these to assess bias in decisions concerning cases that were not considered by the panel. Importantly, the existence of the proposed method could serve as motivation to collect such small sets of gold standard decisions in domains where these are not yet available, with the central goal of assessing biases. As empirically shown in our results, small sets of a few hundred decisions are sufficient for the proposed method to yield high quality results. Thus, small investments to obtain gold standard decisions could yield significant benefits for organizations interested in assessing human decision biases. 

We anticipate that our work will serve as a cornerstone for future research in different directions. In particular, the exploration of productive ways to incorporate decision bias assessments into related research questions and downstream tasks may yield promising research across different disciplines. In machine learning, there may be different ways of utilizing the output of our method when training machine learning algorithms on human-generated labels. We also hope our research will inspire future studies on how to effectively integrate our method into practical applications, helping companies and organizations enhance fairness in their decision-making processes. For instance, should human decision makers be informed that their decisions are biased, and if so, how should this be done? Additionally, what types of human-centered interventions can this method offer to humans as part of strategies aimed at mitigating their biases during labeling or decision-making?  Finally, every new technology will have vulnerabilities. Thus, research on adversarial attacks could explore scenarios where malicious decision makers could intentionally game the proposed algorithm into producing misleading assessments, and develop novel ways of solving the proposed problem. By introducing a comprehensive set of simulations using publicly available data, our hope is that future work can use the same evaluation framework to compare against and surpass the proposed methodology, further advencing the state-of-the-art.

\clearpage
\newpage
\vspace{0.3in}
\linespread{1}
\bibliographystyle{unsrt}
\bibliography{references} 

\clearpage
\begin{appendices}
\renewcommand{\thesection}{\Alph{section}}%
\section{Theoretical Proofs}
\label{section:Proof}

In this appendix, we present the complete proof supporting our theoretical garantee.

\begin{lemma}
Given the correct functional form specification of each human's decision model $f^k: X \mapsto Y'^k$, then $\hat{Y} \independent Y | Y'^k$ and ${Y}'^k \independent Y | \hat{Y}$. 
\end{lemma}

\begin{proof}

Given the correct functional form for $f_k: X_k \mapsto Y_k'$ then $\hat{Y}_k = {Y_k}' + \epsilon$ where the $\epsilon$ is the constant term; and thus, $\hat{Y}_k  Y_k | {Y_k}'$. Analogously, if ${Y_k}' = \hat{Y}_k - \epsilon$, then  ${Y_k}' Y_k |\hat{Y}_k$.
\end{proof}

\addtocounter{theorem}{-2}
\begin{theorem}
Given the correct functional form for the humans' decision models ($f_k: X \rightarrow Y'^k$), then there exists a ratio $\frac{TPR_{\:\hat{Y}|Y', \:A}^l}{PPV_{\:\hat{Y}|Y', \:A}^l} = \frac{TPR_{\:\hat{Y}|Y', \:A}^k}{PPV_{\:\hat{Y}|Y', \:A}^k} = c$, such that if the biases exhibited by a pair of  models of human decisions, $f_l$ and $f_k$, are such that $GAP\:_{\hat{Y}|Y,\:A}^l >
GAP\:_{\hat{Y}|Y,\:A}^k$, then the decision biases of this pair of humans are such that $
GAP\:_{Y'|Y,\:A}^l > GAP\:_{Y'|Y,\:A}^k$. 
\end{theorem}

\begin{proof}

Given $GAP\:_{\hat{Y}|Y,\:A}^l >
GAP\:_{\hat{Y}|Y,\:A}^k$, this can be rewritten as:
\begin{equation}
\label{eq:inquality1}
\begin{split}
    P(\hat{Y_l}= 1 | A = 0, Y = 1) - P(\hat{Y_l}= 1 | A = 1, Y = 1) > \\
    P(\hat{Y_k}= 1 | A = 0, Y = 1) - P(\hat{Y_k}= 1 | A = 1, Y = 1) 
\end{split}
\end{equation}
then
\begin{equation}
\label{eq:inquality2}
\begin{split}
    P(Y_l'= 1 | A = 0, Y = 1) - P(Y_l'= 1 | A = 1, Y = 1) > \\
    P(Y_k'= 1 | A = 0, Y = 1) - P(Y_k'= 1 | A = 1, Y = 1)
    \end{split}
\end{equation}
\newline
It is also true that, 
\begin{equation}\label{eq:proof2eq1}
\begin{split}
P(\hat{Y_i}= 1 | A = a, Y = 1) * P(Y_i' = 1 | A = a,  Y = 1, \hat{Y_i}= 1)= \\ \tfrac{P(\hat{Y_i}= 1, A = a, Y = 1)}{P(A = a, Y = 1)} * \tfrac{P(Y_i' = 1, A = a, Y = 1, \hat{Y_i}= 1)}{P(A = a, Y = 1, \hat{Y_i}= 1)} = \tfrac{P(Y_i' = 1, A = a, Y = 1, \hat{Y_i}= 1)}{P(A = a, Y = 1)} = \\ P(Y_i' = 1, \hat{Y_i}= 1 | A = a, Y = 1)
\end{split}
\end{equation} \newline
By rearranging eq.\ref{eq:proof2eq1}, we have
\begin{equation}
    \begin{split}\label{eq:proof2eq2}
    P(Y_i' = 1, \hat{Y_i}= 1 | A = a, Y = 1) = \\ P(\hat{Y_i}= 1 | A = a, Y = 1)*P(Y_i' = 1| A = a, Y = 1, \hat{Y_i}= 1) 
    \end{split}
\end{equation}
\newline
It is also true that,
\begin{equation}
    \begin{split}\label{eq:proof2eq3}
    \tfrac{P(Y_i' = 1, \hat{Y_i}= 1 | A = a, Y = 1)}{P(Y_i'= 1 | A = a, Y = 1)} =  \tfrac{P(Y_i' = 1, \hat{Y_i}= 1, A = a, Y = 1)}{P(A = a, Y = 1)} * \tfrac{P(A = a, Y = 1)}{P(Y_i'= 1, A = a, Y = 1)} = \\
    P(\hat{Y_i}= 1 | Y_i' = 1, A = a, Y = 1)
    \end{split}
\end{equation}
By rearranging eq.\ref{eq:proof2eq3},
\begin{equation}
\label{eq:proof2eq4}
\begin{split}
    P(Y_i' = 1, \hat{Y_i}= 1 | A = a, Y = 1) = \\
    P(Y_i'= 1 | A = a, Y = 1) * P(\hat{Y_i}= 1 | Y_i' = 1, A = a, Y = 1)
        \end{split}
\end{equation}
From eq.\ref{eq:proof2eq2} and eq.\ref{eq:proof2eq4},
\begin{equation}\label{eq:proof2eq5}
\begin{split}
    P(\hat{Y_i}= 1 | A = a, Y = 1)*P(Y_i' = 1| A = a, Y = 1, \hat{Y_i}= 1) = \\
P(Y_i'= 1 | A = a, Y = 1) * P(\hat{Y_i}= 1 | Y_i' = 1, A = a, Y = 1) 
 \end{split}
\end{equation}
By rearranging eq.\ref{eq:proof2eq5},
\begin{equation}
\label{eq:proof2eq6}
    \tfrac{P(\hat{Y_i}= 1 | A = a, Y = 1)}{P(Y_i'= 1 | A = a, Y = 1)} = \tfrac{P(\hat{Y_i}= 1 | Y_i' = 1, A = a, Y = 1)}{P(Y_i' = 1 | \hat{Y_i}= 1, A = a, Y = 1)}
\end{equation}
\newline
From Lemma, it is true that $\hat{Y} \independent Y | Y'$, and $\hat{Y} \independent Y |A, Y'$; therefore,
\begin{equation}
\label{proof2eq7}
    P(\hat{Y_i}= 1 | A = a, Y_i' = 1) = P(\hat{Y_i}= 1 | Y_i' = 1, A = a , Y = 1)
\end{equation}
and
\begin{equation}
\label{proof2eq8}
    P(Y_i' = 1 | A = a, \hat{Y_i} = 1) = P(Y_i' = 1 | \hat{Y_i} = 1, A = a , Y = 1)
\end{equation}
From eq.\ref{eq:proof2eq6}, eq.\ref{proof2eq7}, and eq.\ref{proof2eq8}, we have
\begin{equation}
    \label{eq:pproof2eq9}
    \tfrac{P(\hat{Y_i}= 1 | A = a, Y = 1) }{P(Y_i'= 1 | A = a, Y = 1)}  = \tfrac{P(\hat{Y_i}= 1 | Y_i' = 1, A = a)}{P(Y_i' = 1 | \hat{Y_i}= 1, A = a)}
\end{equation}
Note that right hand side of eq.\ref{eq:pproof2eq9} is the ``recall (TPR$_{\:\hat{Y}|Y',\:A}$) versus precision (PPV$_{\:\hat{Y}|Y',\:A}$) ratio" and we let the ratio equal to a constant $c$, so
\begin{equation}
    \label{eq:pproof2eq10}
    \tfrac{P(\hat{Y_i}= 1 | Y_i' = 1, A = a)}{P(Y_i' = 1 | \hat{Y_i}= 1, A = a)} = \tfrac{TPR\:_{\hat{Y}|Y', \:A}^i}{PPV\:_{\hat{Y}|Y', \:A}^i} = c
\end{equation}
Note that the ratio c can be simplified as follows:
\begin{equation} 
\label{eq:simplified_constrain2}
   c= \tfrac{TPR_{\:\hat{Y}|Y', \:A}}{PPV_{\:\hat{Y}|Y', \:A}} = \tfrac{\tfrac{TP}{TP + FN}}{\tfrac{TP}{TP + FP}} = \tfrac{TP + FP}{TP + FN} = \tfrac{|\hat{Y}=1, A = a, \sim a|}{|Y'=1, A = a, \sim a|}
\end{equation}
From eq.\ref{eq:pproof2eq9} and eq.\ref{eq:pproof2eq10}, it is true that
\begin{equation}
\label{eq:pproof2eq11}
    \tfrac{P(\hat{Y_i}= 1 | A = a, Y = 1)}{c} = P(Y_i' = 1 | A = a, Y = 1)
\end{equation}
Given eq.\ref{eq:inquality1} above: 
\begin{equation}
\begin{split}
    P(\hat{Y_l}= 1 | A = 0, Y = 1) - P(\hat{Y_l}= 1 | A = 1, Y = 1) > \\
    P(\hat{Y_k}= 1 | A = 0, Y = 1) - P(\hat{Y_k}= 1 | A = 1, Y = 1) 
\end{split}
    \tag{\ref{eq:inquality1}}
\end{equation}
dividing both sides by $c$, we have:
\begin{equation}
\label{eq:inquality3}
\begin{split}
    \frac{P(\hat{Y_l}= 1 | A = 0, Y = 1)}{c} - \frac{P(\hat{Y_l}= 1 | A = 1, Y = 1)}{c} > \\
    \frac{P(\hat{Y_k}= 1 | A = 0, Y = 1)}{c} - \frac{P(\hat{Y_k}= 1 | A = 1, Y = 1)}{c}
\end{split}
\end{equation}
which is equivalent to
\begin{equation}
\begin{split}
    P(Y_l'= 1 | A = 0, Y = 1) - P(Y_l'= 1 | A = 1, Y = 1) > \\
    P(Y_k'= 1 | A = 0, Y = 1) - P(Y_k'= 1 | A = 1, Y = 1)
\end{split}
    \tag{\ref{eq:inquality2}}
\end{equation}

\end{proof}

\begin{theorem}
Given the correct functional form for the humans' decision models ($f: X \rightarrow Y'$), then there exits a ratio $\frac{TPR_{\:\hat{Y}|Y', \:A}}{PPV_{\:\hat{Y}|Y', \:A}} = c$ such that the bias in the human's model, $GAP\:_{\hat{Y}|Y,\:A}$, and decision bias of the human, $GAP\:_{Y'|Y,\:A}$, follow the relationship: $\frac{GAP\:_{\hat{Y}|Y,\:A}}{c} = GAP\:_{Y'|Y,\:A}$, where $
GAP\:_{\hat{Y}|Y,\:A} = TPR\:_{\hat{Y}|Y, \:a} - TPR\:_{\hat{Y}|Y,\:\sim a}$
and $GAP\:_{Y'|Y,\:A} = TPR\:_{Y'|Y, \:a} - TPR\:_{Y'|Y,\:\sim a}$.
\end{theorem}

\begin{proof}
    From Lemma and Theorem \ref{theorem}, we have
\begin{equation}
     \tag{\ref{eq:pproof2eq10}}
    \tfrac{P(\hat{Y_i}= 1 | Y_i' = 1, A = a)}{P(Y_i' = 1 | \hat{Y_i}= 1, A = a)} = \tfrac{TPR\:_{\hat{Y}|Y', \:A}^i}{PPV\:_{\hat{Y}|Y', \:A}^i} = c
\end{equation}
Therefore, 
\begin{equation}
\label{theorem_eq1}
 \frac{GAP\:_{\hat{Y}|Y,\:A}}{c} = GAP\:_{Y'|Y,\:A}
 \end{equation}
\end{proof}

\section{Comparison between MDBA and MDBA-Naive}
\label{sec:ablation_study}

Tables \ref{tab: comparison-mdba-Naive-easyForSR} and \ref{tab: comparison-mdba-Naive-hardForSR} show the results of the comparison between MDBA and MDBA-Naive for {\itshape correct} and {\itshape incorrect} within-group orderings, respectively.  MDBA-Naive follows most of the key steps but omits Step 2 in Figure \ref{fig:method_key_ideas}. Specifically, after training a model for each human ($f^k$ for $H^k$), we skip the bias mitigation step (line 3 in Algorithm \ref{alg:mba_version3}) and directly apply the models to evaluate the disjoint and scarce gold standard dataset ($GS$) without adjusting the thresholds to satisfy c values.

The results indicate that in some settings, the inclusion of the bias mitigation strategy is significantly beneficial. In a few cases, MDBA-Naive slightly outperforms MDBA, but no statistical significance was found, suggesting they are comparable. Overall, MDBA consistently performs better or is comparable to MDBA-Naive, underscoring that Step 2, the bias mitigation strategy, is crucial in most cases and, at the very least, not harmful.

\setcounter{table}{0}
\renewcommand{\thetable}{B\arabic{table}}
\begin{table}[htbp]
\pgfplotstableset{highlight/.append style={
    postproc cell content/.append code={
            \pgfkeysalso{@cell content=\textbf{##1}}%
    }
}}
  \begin{center}
  \singlespacing
\begin{threeparttable}

    \caption{MDBA and MDBA-Naive (theoretical garantee removed) performance measured by MAE for humans exhibiting {\itshape correct} within-group orderings }
    \label{tab: comparison-mdba-Naive-easyForSR}
    \sisetup{
        round-mode          = places, 
        round-precision     = 2, 
    }
    \begin{filecontents*}{EasyForSR_Ablations.csv}
Dataset,GT Per Group,MDBA,Variant,MDBA_improv,MDBA,Variant,MDBA_improv
Adult,100,0.082,0.099,17.1\%,0.071,0.076,7.4\%
,200,0.073,0.08,9.1\%,0.043,0.05,14.5\%*
,300,0.076,0.079,2.9\%,0.044,0.052,15.4\%
,400,0.066,\textbf{0.066},0.3\%,0.036,0.043,14.6\%
Credit,100,0.129,0.184,29.8\%***,0.097,0.121,19.4\%*
,200,0.099,0.16,38.2\%***,0.077,0.105,26.7\%**
,300,0.095,0.155,38.6\%***,0.074,0.103,28\%***
,400,0.089,0.161,44.5\%***,0.073,0.101,27.7\%***
Employee,100,0.082,0.109,24.7\%**,0.072,0.093,22.4\%*
,200,0.057,0.076,25.5\%***,0.06,0.08,25.1\%*
,300,0.045,0.069,34.8\%***,0.054,0.077,30.8\%**
,400,0.046,0.064,28.1\%**,0.049,0.07,30.3\%**
Readmission,100,0.081,0.103,21.2\%,0.078,0.083,6.3\%
,200,0.059,0.089,34\%***,0.04,0.042,3.7\%
,300,0.052,0.07,26.1\%*,0.035,0.036,1.6\%
,400,0.05,0.061,19.4\%,0.029,0.03,4.2\%
\end{filecontents*}
\pgfplotstabletypeset[
    font=\footnotesize,
      multicolumn names, 
      col sep=comma, 
      column type=l,
      display columns/0/.style={string type ={ S},column name={Dataset}, column type=c|},  
      display columns/1/.style={string type ={S},column name={\textsc{GS Per}}, column type=c|},
      display columns/2/.style={highlight, string type ={S},column name={MDBA}, column type=c|},
      display columns/3/.style={string type ={S},column name={MDBA-}},
      display columns/4/.style={string type ={S},column name={MDBA}, column type=c|},
      display columns/5/.style={highlight, string type ={S},column name={MDBA}, column type=c|},
      display columns/6/.style={string type ={S},column name={MDBA-}},
      display columns/7/.style={string type ={S},column name={MDBA}, column type=c|},
      every head row/.style={
		before row={\toprule
		&  & \multicolumn{3}{c|}{20\% Positive Label Prevalence} & \multicolumn{3}{c|}{30\% Positive Label Prevalence}\\}, 
		after row={& \textsc{worker} & & Naive &IMPROV. & & Naive & IMPROV.\\
			\midrule } 
			},
		every last row/.style={after row=\bottomrule}, 
            every row no 4/.style={before row=\midrule},
		every row no 8/.style={before row=\midrule},
		every row no 12/.style={before row=\midrule}
    ]{EasyForSR_Ablations.csv} 
    
      \begin{tablenotes}\scriptsize
  \linespread{1}
           \item\hspace*{-\fontdimen2\font}humans' bias estimation errors. Values show Mean Absolute Error (MAE). MDBA IMPROV. shows the improvement of MDBA over the MDBA-Naive.  MDBA yields substantially better and otherwise comparable estimations of workers accuracies. *** MDBA is statistically significantly better ($p<0.01$), **: ($p<0.05$), and *: ($p<0.1$).
    \end{tablenotes}
    \end{threeparttable}
      \end{center}
\end{table}

\begin{table}[htbp]
\pgfplotstableset{highlight/.append style={
    postproc cell content/.append code={
            \pgfkeysalso{@cell content=\textbf{##1}}%
    }
}}
  \begin{center}
\begin{threeparttable}
\singlespacing
    \caption{MDBA and MDBA-Naive (theoretical garantee removed) performance measured by MAE for humans exhibiting {\itshape incorrect} within-group orderings }
    \label{tab: comparison-mdba-Naive-hardForSR}
    \sisetup{
        round-mode          = places, 
        round-precision     = 2, 
    }
    \begin{filecontents*}{HardForSR_Ablations.csv}
Dataset,GT Per Group,MDBA,Variant,MDBA_improv,MDBA,Variant,MDBA_improv
Adult,100,0.069,0.077,10.2\%,0.069,0.08,13.8\%
,200,0.056,0.061,7.9\%,0.055,0.061,10.4\%
,300,0.051,0.053,4.5\%,0.044,0.049,11\%
,400,\textnormal{0.06},\textbf{0.058},-2\%,0.039,0.044,12\%
Credit,100,0.097,0.116,16.2\%,0.071,0.072,1.8\%
,200,0.082,0.099,17.1\%,\textnormal{0.065},\textbf{0.059},-9.7\%
,300,0.065,0.089,27.1\%**,\textnormal{0.056},\textbf{0.052},-7.8\%
,400,0.067,0.091,25.8\%**,\textnormal{0.059},\textbf{0.054},-8.3\%
Employee,100,0.086,0.125,31.3\%***,0.102,0.116,11.7\%
,200,0.068,0.1,32.2\%***,0.075,0.089,15.6\%
,300,0.056,0.087,36.4\%***,0.071,0.087,19.3\%**
,400,0.053,0.078,32.6\%**,0.067,0.081,17.5\%**
Readmission,100,0.09,0.116,22.9\%**,0.051,0.058,12.2\%
,200,0.051,0.072,28.4\%***,0.041,0.044,7.8\%
,300,0.047,0.058,19.3\%*,0.029,0.033,14.1\%
,400,0.041,0.055,25.4\%**,0.031,0.035,9.9\%
\end{filecontents*}
\pgfplotstabletypeset[
    font=\footnotesize,
      multicolumn names, 
      col sep=comma, 
      column type=l,
      display columns/0/.style={string type ={ S},column name={Dataset}, column type=c|},  
      display columns/1/.style={string type ={S},column name={\textsc{GS Per}}, column type=c|},
      display columns/2/.style={highlight, string type ={S},column name={MDBA}, column type=c|},
      display columns/3/.style={string type ={S},column name={MDBA-}},
      display columns/4/.style={string type ={S},column name={MDBA}, column type=c|},
      display columns/5/.style={highlight, string type ={S},column name={MDBA}, column type=c|},
      display columns/6/.style={string type ={S},column name={MDBA-}},
      display columns/7/.style={string type ={S},column name={MDBA}, column type=c|},
      every head row/.style={
		before row={\toprule
		&  & \multicolumn{3}{c|}{20\% Positive Label Prevalence} & \multicolumn{3}{c|}{30\% Positive Label Prevalence}\\}, 
		after row={& \textsc{worker} & & Naive &IMPROV. & & Naive & IMPROV.\\
			\midrule } 
			},
		every last row/.style={after row=\bottomrule}, 
            every row no 4/.style={before row=\midrule},
		every row no 8/.style={before row=\midrule},
		every row no 12/.style={before row=\midrule}
    ]{HardForSR_Ablations.csv} 
    
      \begin{tablenotes}\scriptsize
  \linespread{1}
           \item\hspace*{-\fontdimen2\font}humans' bias estimation errors. Values show Mean Absolute Error (MAE). MDBA IMPROV. shows the improvement of MDBA over the MDBA-Naive.  MDBA yields substantially better and otherwise comparable estimations of workers accuracies. *** MDBA is statistically significantly better ($p<0.01$), **: ($p<0.05$), and *: ($p<0.1$). Negative MDBA IMPROVs indicate that the MDBA-Naive is comparable to MDBA, with no statistically significant difference found between them.
    \end{tablenotes}
    \end{threeparttable}
      \end{center}
\end{table}

\clearpage
\section{Pseudocode for Simulating Biased Human Decision}
\label{sec:simulation_code}
In this appendix, we provide the pseudocode for simulating biased human decisions for the sensitive group (i.e., $A = a$): \emph{correct within-group ordering} (Algorithm \ref{alg:simulation_correct_orderings}) and \emph{incorrect within-group ordering} (Algorithm \ref{alg:simulation_incorrect_orderings}), as described in Section \ref{sec:empiric}.
\begin{algorithm2e}[!htb]
\caption{{\itshape Correct} Within-group Orderings Simulation}
\LinesNumbered
\SetAlFnt{\small\sf}
\label{alg:simulation_correct_orderings}
\DontPrintSemicolon  

\SetKwFunction{FMBA}{Correct Within-group Orderings}
\SetKwProg{Fn}{Simulation}{:}{}
\Fn{\FMBA{$\{X^k, Y^k\}_{k=1}^K$, $\{tpr^k\}_{k=1}^K$}}{
    \BlankLine
    \ForEach{$\{X^k, Y^k\} \in \{X^k, Y^k\}_{k=1}^K$}{
        $f_{sim}^k \gets$ Train a model on  $\{X^k, Y^k\}$ \;
        $Prob^k \gets$ Obtain predicted probabilities using $f_{sim}^k$ on $Y^k$ \;
        
        \ForEach{$Prob^k_i \in \{Prob^k_i\}_{i=1}^{n^k}$}{ 
            $Prob_t \gets$  $Prob^k_i$ \;
            \ForEach{$Prob^k_i \in \{Prob^k_i\}_{i=1}^{n^k}$}{
                ${Y'}_{i,a}^k \gets$ 1 \textbf{if} $Prob^k_i \geq Prob_t$ \textbf{else} 0 \;
            }
            
            \lIf{$TPR_{Y'^k|Y^k,a} \in [tpr^k \pm 0.01]$}{
                $Prob_{opt}^k \gets Prob_t$
            }
        }
        
        \textbf{if} multiple or no $Prob_{opt}^k$ is found \textbf{then} $Prob_{opt}^k \gets$ $Prob_t$ corresponding to the $TPR_{Y'^k|Y^k,a}$ that is closest to $tpr^k$ \;

        \ForEach{$Prob^k_i \in \{Prob^k_i\}_{i=1}^{n^k}$}{
            ${Y'}_{i,a}^k \gets$ 1 \textbf{if} $Prob^k_i \geq Prob_{opt}^k$ \textbf{else} 0 \;
        }
    }
    
    \KwRet $\{{Y'}_a^k\}_{k=1}^K$
}
\end{algorithm2e}

\begin{algorithm2e}[!htb]
\caption{{\itshape Incorrect} Within-group Orderings Simulation}
\LinesNumbered
\SetAlFnt{\small\sf}
\label{alg:simulation_incorrect_orderings}
\DontPrintSemicolon  
\SetKwRepeat{Do}{do}{while}
\SetKwFunction{FMBA}{Incorrect Within-group Orderings}
\SetKwProg{Fn}{Simulation}{:}{}

\Fn{\FMBA{$\{X^k, Y^k\}_{k=1}^K$, $tpr_{low}$, $tpr_{gap}$, $GS = \{X_l, Y_l\}_{l=1}^m$, $p$, $Step$}}{
    \BlankLine
    $T \gets GS$ \;
    \ForEach{$\{X^k, Y^k\} \in \{X^k, Y^k\}_{k=1}^K$}{
        $T \gets T \cup \{X^k, Y^k\}$ \;
    }
$f_{sim} \gets$ Train a model on $T$ \;
$Coef \gets$ Extract coefficients from $f_{sim}$ \;
    
    \ForEach{$\{X^k, Y^k\} \in \{X^k, Y^k\}_{k=1}^K$}{
        $Coef_{temp} \gets Coef$ \;
        $Coef_{opt} \gets []$ \;

        \Do{$Coef_{opt} = []$}{
            $Prob^k \gets$ Obtain predicted probabilities by applying $f_{sim}$ with $Coef_{temp}$ on $X_a^k$ \;
            \ForEach{$Prob_i^k \in Prob^k$}{
                ${Y'}_{i,a}^k \gets$ 1 \textbf{if} $Prob_i^k$ is among the top $p$ in $Prob^k$, \textbf{else} 0 \;
            }
       
            \lIf{$k = 1$ \textbf{and} $TPR_{Y'^k|Y^k,a} \in [tpr_{low} \pm tpr_{gap}]$}{
                $Coef_{opt} \gets Coef_{temp}$ \textbf{break}
            }
            
            \lIf{$k \neq 1$ \textbf{and} $TPR_{Y'^k|Y^k,a} \geq tpr^{k-1} + tpr_{gap}$}{
                $Coef_{opt} \gets Coef_{temp}$ \textbf{break}
            }
            
            \Else{
                $Coef_{temp} \gets Coef_{temp} - Step$
            }
        }
        
        $Prob^k \gets$ Obtain predicted probabilities by applying $f_{sim}$ with $Coef_{opt}$ on $X^k$ \;
        
        \ForEach{$Prob_i^k \in Prob^k$}{
            ${Y'}_{i,a}^k \gets$ 1 \textbf{if} $Prob_i^k$ is among the top $p$ in $Prob^k$, \textbf{else} 0 \;
        }
    }
    
    \KwRet $\{{Y'}_a^k\}_{k=1}^K$
}
\end{algorithm2e}

\end{appendices}

\end{document}